\documentclass[runningheads]{llncs}

\usepackage{eccv}

\usepackage{eccvabbrv}

\usepackage{url}            
\usepackage{booktabs}       
\usepackage{amsfonts}       
\usepackage{nicefrac}       
\usepackage{microtype}      
\usepackage{xcolor}         
\usepackage{graphicx}
\usepackage{enumitem}
\usepackage{amsmath,bm}
\usepackage{multirow}

\usepackage[accsupp]{axessibility}  

\usepackage[pagebackref,breaklinks,colorlinks,citecolor=eccvblue]{hyperref}

\usepackage[pagebackref,breaklinks,colorlinks,citecolor=eccvblue]{hyperref}

\usepackage{orcidlink}

\newcommand{\nickname}{SIR}

\begin{document}

\title{\nickname{}: Multi-view Inverse Rendering with Decomposable Shadow Under Indoor Intense
Lighting} 

\author{Xiaokang Wei$^{1,2}$, Zhuoman Liu$^{1}$, Ping Li$^{1}$, Yan Luximon$^{1,2}$\thanks{Corresponding author}} 
\institute{$^{1}$The Hong Kong Polytechnic University,\\
$^{2}$Laboratory for Artificial Intelligence in Design, HKSAR\\
\email{\{xiaokang.wei,zhuo-man.liu\}@connect.polyu.hk, 
\{p.li, yan.luximon\}@polyu.edu.hk}}
%
\authorrunning{X.Wei, Z.Liu, Ping Li and Y.Luximon}
\titlerunning{Shadow Inverse Rendering}
\maketitle

\begin{abstract}
We propose \nickname{}, an efficient method to decompose differentiable shadows for inverse rendering on indoor scenes using multi-view data, addressing the challenges in accurately decomposing the materials and lighting conditions. Unlike previous methods that struggle with shadow fidelity in complex lighting environments, our approach explicitly learns shadows for enhanced realism in material estimation under unknown light positions. Utilizing posed HDR images as input, \nickname{} employs an SDF-based neural radiance field for comprehensive scene representation. Then, \nickname{} integrates a shadow term with a three-stage material estimation approach to improve SVBRDF quality. Specifically, \nickname{} is designed to learn a differentiable shadow, complemented by BRDF regularization, to optimize inverse rendering accuracy. Extensive experiments on both synthetic and real-world indoor scenes demonstrate the superior performance of \nickname{} over existing methods in both quantitative metrics and qualitative analysis. The significant decomposing ability of \nickname{} enables sophisticated editing capabilities like free-view relighting, object insertion, and material replacement. The code and data are available at \href{https://xiaokangwei.github.io/SIR/}{https://xiaokangwei.github.io/SIR/}.

\end{abstract}

\section{Introduction}

\begin{figure}[t]
\centering
  \includegraphics[width=0.99\linewidth]{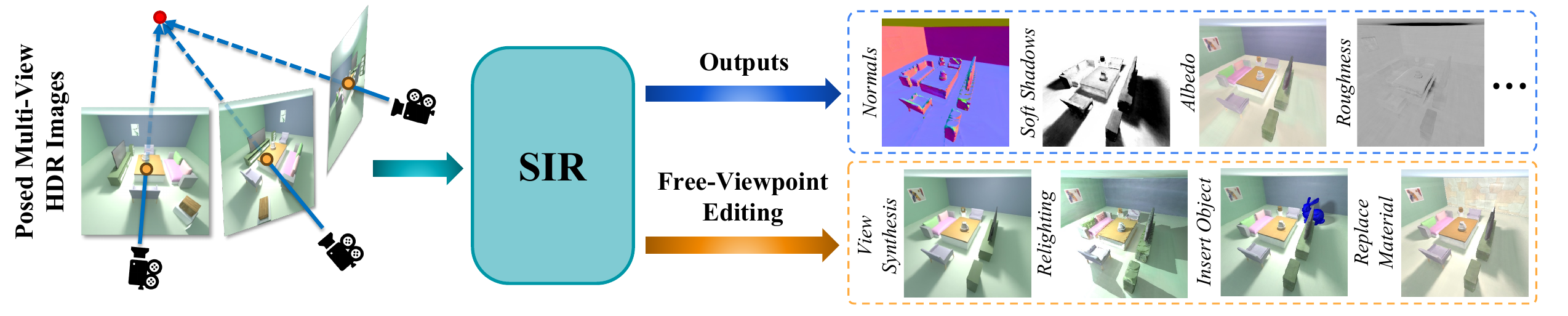}
  \vspace{-0.4cm}
\caption{Given posed multi-view HDR images of an indoor scene, \textbf{\nickname{}} successfully disentangle the scene appearance into inherent attributes, which can produce convincing results for several free-viewpoint editing applications.}
\label{fig:intro}
\vspace{-0.6cm}
\end{figure}

Inverse rendering, a crucial concept in computer graphics and vision, involves deducing scene properties such as geometry, lighting, and materials from images. This process, which inverts the traditional rendering pipeline, has broad applications~\cite{guo2019,meka2018,philip2019,meka2020} ranging from photo-realistic editing to augmented reality. The objective is to achieve a high level of realism and accuracy, facilitating a flawless blend of existing scene elements with newly introduced components or properties. 
However, the complexity inherent in scenes, especially regarding lighting and material attributes, poses significant challenges. 

The first challenge lies in the representation of scene geometry and lighting. \textbf{i)} For geometry, ranging from conventional explicit representations like point clouds and meshes to recent developments in implicit representations, such as neural radiance fields (NeRFs)~\cite{mildenhall2021} and DeepSDF~\cite{park2019}, vary in their efficiency and computational demands. The fidelity of the yielded scene geometry directly influences the success of inverse rendering approaches. \textbf{ii)} For lighting,  the choice between low dynamic range (LDR)~\cite{wang2021learning,li2022physically} and high dynamic range (HDR)~\cite{srinivasan2020,haefner2021recovering} fields significantly affects the outcome of the rendering. The former is more computationally efficient, while the latter enables accurate measurements of scene radiance from a collection of photographs with varying exposures~\cite{debevec2008}. 

The second challenge lies in material estimation, specifically in the decomposition of shadow and albedo. Shadows are highly prevalent and noticeable in indoor scenes because of the primary light sources and the intricate geometric occlusions between objects. This necessitates the decomposition of shadows to achieve realistic effects in scene editing applications.
Existing multi-view methods for indoor scenes have not yet effectively resolved this problem. ~\cite{nvdiffrec} tends to bake pre-existing shadows into the learned albedo results. Such deprecated shadows in their albedo results can lead to inaccurate rendering during editing. For example, if the scene is illuminated by a light source placed at a different position, the shadows they produce cannot shift accordingly as expected.
On the other hand, ~\cite{li2022physically} proposes a single-view inverse rendering method to effectively decompose shadows. However, it relies on supervision with direct shading, excluding shadow effects, which are not available in real-world scenes. ~\cite{li2022neural} solves shadow decomposition through self-supervision but cannot handle unknown light source positions. Consequently, accurately estimating shadows without explicit supervision under unknown light sources remains a significant challenge.

To address these challenges, we propose \textbf{\nickname{}} (\textbf{S}hadow \textbf{I}nverse \textbf{R}endering) (see Fig.~\ref{fig:intro}), which efficiently decomposes shadows for indoor scenes without explicit supervision. \nickname{} employs an SDF-based neural radiance field for scene representation and leverages multi-view HDR images to provide more information for lighting.
Then, for unsupervised shadow decomposition, we propose a three-stage material estimation, including \textit{albedo initialization with hard shadows}, \textit{albedo refinement with differentiable soft shadows}, and \textit{roughness refinement}. We introduce instance-level regularization to refine BRDF attributes (albedo and roughness).
This approach not only elevates SVBRDF quality but also facilitates editing capabilities. In summary, the main contributions of this work include: 

\vspace{-0.2cm}
\begin{itemize}[leftmargin=*]
    \item We propose SIR, a new inverse rendering framework with a novel shadow term to decompose shadow and albedo in multi-view indoor scenes effectively. 
    \item We present a three-stage material estimation strategy, incorporating differentiable shadow and BRDF regularization.
    \item Our method exhibits superior inverse rendering accuracy over current methods on both synthetic and real-world indoor scene datasets. Additionally, its robustness is further demonstrated through successful scene editing.
\end{itemize}

\section{Related Works}

\subsection{Neural Scene Representations}


Recent progress in neural scene representations has enhanced the recovery of scene geometry and radiance for inverse rendering tasks. Neural radiance field (NeRF)~\cite{mildenhall2021} employs a single Multilayer Perceptron (MLP) to process a 5D light field, thereby learning the radiance field for generating new viewpoints. However, the geometric accuracy of NeRF for inverse rendering is constrained due to the inherent ambiguity in volume rendering~\cite{zhang2020}. To address this, methods like NeuS~\cite{wang2021} and VolSDF~\cite{yariv2021} have been developed, transforming Signed Distance Functions (SDF)~\cite{park2019} into densities for more accurate volume rendering.
In addition to these implicit approaches, explicit representation methods~\cite{liao2018,chen2021} have also been proposed to accelerate the rendering process. Among them, DMTet~\cite{shen2021} and other DMTet-based solutions~\cite{nvdiffrec,nvdiffrecmc} utilize a differentiable marching tetrahedral layer to refine surface meshes.
Recently, many neural scene representations have been proposed with hybrid methods such as InstantNGP~\cite{muller2022instant} and 3D Gaussian splatting~\cite{kerbl20233d}, which significantly improve the efficiency and reduce the computational cost. However, these methods often struggle to accurately capture the complex topologies and small-scale structures in indoor scenes. To effectively reconstruct the geometry of indoor scenes, we propose an optimization of topology, lighting, and materials using an SDF-based neural radiance field.

\vspace{-0.2cm}
\subsection{Lighting Estimation}


For inverse rendering, accurately estimating lighting, particularly in indoor scenes, is a complex and essential task. Most current illumination estimation methods operate on single images, with a primary emphasis on integrating virtual objects into real images rather than making substantial alterations to the scene's illumination~\cite{karsch2011rendering,garon2019fast,zhan2021emlight}. While traditional methods like a single environment map~\cite{gardner2017,legendre2019} and spherical lobes~\cite{garon2019} have been used, they often neglect spatial variations and high-frequency details in lighting. Recent innovations~\cite{song2019,srinivasan2020} have attempted to improve 3D lighting representation, but still grapple with challenges like spatial instability and the lack of HDR information. ~\cite{li2020} propose per-pixel spatially-varying spherical Gaussians (SVSG) lighting representation to capture high-frequency effects and demonstrate that SGs are superior to spherical harmonics (SH) for depicting lighting details in indoor scenes. Neural-PIL~\cite{boss2021} proposes a pre-integrated lighting network based on image-based lighting (IBL), showing better performances on conveying global illumination than SGs and SH. Hence, we utilize a neural HDR-radiance field to represent the IBL at any spatial point, thereby ensuring a more accurate and detailed depiction of indoor lighting scenarios with a focus on physically accurate HDR lighting prediction.

\vspace{-0.2cm}
\subsection{Material Estimation}


Material estimation in inverse rendering can be categorized into two levels: object level and scene level. Object-level estimation~\cite{physg,invrender,nvdiffrec,liang2022spidr,boss2021nerd} focuses on individual objects, often in controlled or simplified environments. Object-level approaches are generally less complex, as they deal with fewer variables and more straightforward lighting conditions. In contrast, scene-level material estimation~\cite{iblnerf,textir} is significantly more challenging due to the complexity and variability of entire scenes. This includes diverse lighting, multiple objects with different materials, and shadows.

The complexity of scene-level material estimation is further compounded by the choice between single-view and multi-view approaches. Single-view material estimation~\cite{li2020,gardner2017}, despite its simplicity and lower data requirements, often faces the ill-posed issue, where insufficient information leads to ambiguous or inaccurate estimations. This is particularly evident in complex scenes where a single viewpoint cannot capture the entirety of the scene's lighting and material properties.
In contrast, multi-view material estimation~\cite{iblnerf,invrender,physg,nvdiffrec} leverages images from multiple viewpoints, providing a more comprehensive understanding of the scene. It can significantly reduce the ambiguity associated with single-view estimations, allowing for more accurate and reliable material property extraction. Our work utilizes multi-view images for material estimation, specifically addressing the challenges at the scene level.


\section{Methods}

\subsection{Overview}
\begin{figure*}[t]
\centering
  \includegraphics[width=0.99\linewidth]{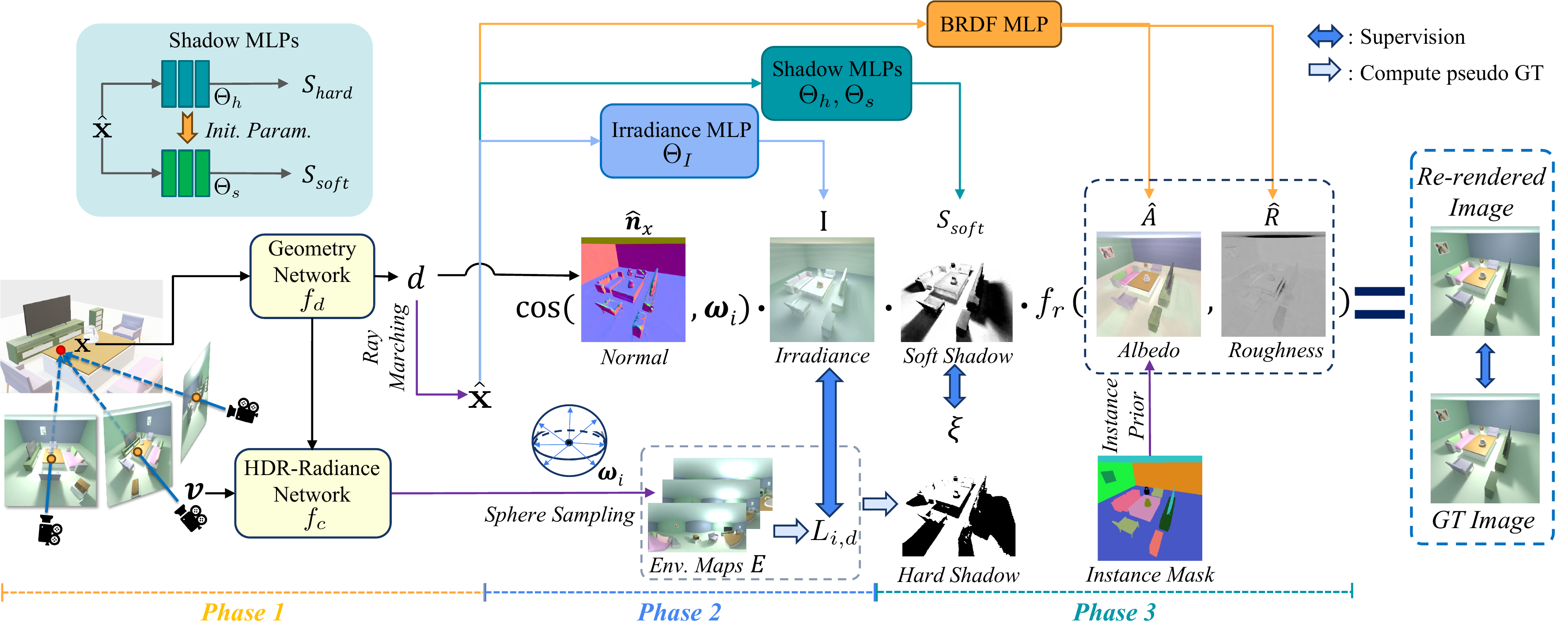}
  \vspace{-0.4cm}
\caption{
The pipeline consists of three phases: \textbf{1)} We sample a ray with direction $\bm{v}$ and spatial point $\textbf{x}$ from the given posed HDR images. The geometry network $f_d$ learns the signed distance $d$ to obtain surface point $\hat{\textbf{x}}$, and the HDR-radiance network $f_c$ learns radiance $\hat{C}$. \textbf{2)} We obtain the diffuse incoming lighting $L_{i,d}$ by integrating incident radiance from environment maps $E$ for learning irradiance $I$. \textbf{3)} Hard shadow $S_{hard}$ is learned using $\Theta_h$ with pseudo ground truth $\xi$. We then initialize the parameters of $\Theta_s$ using the optimized parameters of $\Theta_h$. Instance-level BRDF regularizers are applied, and the whole rendering equation is optimized to update $\hat{A}$, $\hat{R}$, and $S_{soft}$.}
\label{fig:overview}
\vspace{-0.6cm}
\end{figure*}

Given a set of posed HDR images of an indoor scene, our method aims to accurately recover the geometry, global illumination, shadows, and spatially-varying bidirectional reflectance distribution functions (SVBRDFs) (see Fig.~\ref{fig:overview}), which can be used in editing applications with accurate shadow. We use the SDF to represent global illumination and geometry as an HDR-radiance field, and the additional MLPs to represent an irradiance field (Sec.~\ref{sec:lighting}). To handle the ambiguity between shadows and albedo well, we introduce a hard shadow term based on the HDR neural radiance field and propose a shadow field to represent the continuous shadows (Sec.~\ref{sec:shadow}). To eliminate residual shadow and efficiently improve the quality of SVBRDFs, we propose a three-stage material estimation strategy with differentiable shadows and BRDF regularization (Sec.~\ref{sec:brdf}).

\subsection{Background}
The traditional rendering equation~\cite{kajiya1986} is formulated following the principles of energy conservation in physics. It computes the outgoing radiance $L_o$ at surface point $\hat{\textbf{x}}$ with surface normal $\hat{\bm{n}}_x$ along view direction $\bm{\omega}_o$ by integrating over the hemisphere $\Omega^+=\{\bm{\omega}_i:\bm{\omega}_i\cdot\hat{\bm{n}}_x>0\}$, where $\bm{\omega}_i$ is the incident light direction:
\begin{equation}
    L_{o}(\hat{\textbf{x}}, \bm{\omega}_o) = \int_{\Omega^+}L_{i}(\hat{\textbf{x}},\bm{\omega}_i)f_r(\hat{\textbf{x}},\bm{\omega}_i,\bm{\omega}_o)(\bm{\omega}_i\cdot\hat{\bm{n}}_x)\text{d}\bm{\omega}_i \label{con:renderequation}
\end{equation}
\noindent The function $L_{i}(\hat{\textbf{x}},\bm{\omega}_i)$ represents the incoming radiance at the surface point $\hat{\textbf{x}}$ from $\bm{\omega}_i$, while the BRDF function $f_r$ quantifies the proportion of light arriving from direction $\bm{\omega}_i$ that is reflected in the direction $\bm{\omega}_o$ at point $\hat{\textbf{x}}$.

\subsection{Geometry and Lighting Representaion}\label{sec:lighting}
Guided by the rendering equation, physically-based shading models rely on high-quality reconstructed geometry. However, density-based NeRF~\cite{mildenhall2021} struggles to recover satisfactory geometries. Motivated by the effectiveness of SDF-based NeRF~\cite{yariv2021,wang2021}, our method employs VolSDF~\cite{yariv2021} as the backbone to represent geometry and radiance field. It is noted that alternative SDF-based NeRF variants are also applicable~\cite{fu2022geo,li2023neuralangelo,yariv2023bakedsdf}. In particular, we replace the LDR input of VolSDF with HDR images because HDR contains richer environmental lighting information. The attributes of a 3D scene, including SDF and HDR-radiance, are parameterized by a \textit{geometry network} \(\textit{f}_{d}\) and an \textit{HDR-radiance network} \(\textit{f}_{c}\) by:
\begin{equation}
  d = f_{d}(\textbf{x}), \quad \textbf{c} = f_{c}(\textbf{x}, \bm{v})
\end{equation}
\noindent where the geometry network \(\textit{f}_d\) maps a spatial position \(\textbf{x}\) to its signed distance to the surface and the HDR-radiance network $\textit{f}_c$ predict HDR radiance value conditioned on position $\textbf{x}$ and view direction \(\bm{v}\). To render a pixel, we utilize volume rendering similar to NeRF~\cite{mildenhall2021}, which accumulates the density \(\sigma\) from SDF and radiance \(\textbf{c}\) along the view direction \(\bm{v}\) of ray \textbf{r} by:
\begin{equation}
    \hat{C}(\textbf{r}) = \sum_{i=1}^{P} T_i \alpha_{i} \textbf{c}_{i} \label{volumerendering}
\end{equation}
\noindent where \(T_i = \exp(-\sum_{j=1}^{i-1}\sigma_j \delta_j)\) and \(\alpha_i = 1 - \exp(-\sigma_i \delta_i)\) denote the transmittance and alpha value of sampled point, respectively. \(\delta_i\) is the distance between neighboring sampled points. \(P\) is the number of sampled points along a ray. \(\sigma_i\) and \(\textbf{c}_i\) are predicted by the HDR-radiance network $f_c$.
Following VolSDF~\cite{yariv2021}, we convert the SDF value $\textit{d}$ to density $\sigma$ and obtain high-quality smooth surface from the zero level-set of predicted SDF in space \( \Omega \in \mathbb{R}^3 \), where the surface normal is computed as the gradient of SDF w.r.t. surface point $\hat{\textbf{x}}$, \ie, $\hat{\bm{n}}_x = \frac{\nabla {d}_{\Omega}(\hat{\textbf{x}})}{\|\nabla {d}_{\Omega}(\hat{\textbf{x}})\|}$.

We adopt image-based lighting to represent scene environment lighting, which can deal with direct and indirect lighting conditions well. Inspired by InvRender~\cite{invrender}, we obtain spatially-varying environment map \(\textit{E} \in \mathbb{R}^{3 \times H \times W} \) of an indoor scene derived from the pre-trained HDR-radiance field. By querying the environment map \(\textit{E}\), the diffuse incoming lighting $L_{i,d}$ at position \(\textbf{x}\) can be calculated as the integration of incident radiance along any \(\bm{\omega_i}\) over hemisphere \(\Omega^+\):
\begin{equation}
    L_{i,d} (\textbf{x}) = \int_{\Omega^+} E(\textbf{x},\bm{\omega}_i) (\bm{\omega}_i \cdot \bm{n}_x) \text{d}\bm{\omega}_i, 
\end{equation}
\begin{equation}
E(\textbf{x},\bm{\omega}_i) = \hat{C}(\textbf{p}_x, -\bm{\omega}_i)
\end{equation}

\noindent where \(\bm{\omega_i}\) is the incident radiance direction,  \(\textbf{p}_x\) is the intersection position with scene surface along a ray starting from $\textbf{x}$ with a direction $-\bm{\omega}_i$ by ray tracing. Here, we use a simple convolution operation on the environment map \(\textit{E}\) to calculate diffuse incoming lighting $L_{i,d}$ at any spatial position \textbf{x}.

To capture a continuous and spatially varied lighting environment, we further introduce an \textit{irradiance field} \(\Theta_{I}(\textbf{x})\) to learn the irradiance ${I}$ at any input 3D location \(\textbf{x}\), supervised by pre-computed incoming lighting $L_{i,d}$. Instead of restricting to surface points, $\Theta_I$ utilizes arbitrary spatial positions as inputs, efficiently modeling a compact and continuous spatially varying and global illumination of indoor scenes.
It can also significantly enhance the fidelity of the lighting model in complex spatial applications.




\subsection{Hard Shadow Estimation}\label{sec:shadow}
Shadow depicts the occlusion between geometry and ambient lighting. The phenomenon of cast shadow is significant in indoor scenes with relatively strong primary light sources. We note that gradient optimization during inverse rendering tends to bake shadows into the albedo~\cite{nvdiffrecmc}. To enable high-quality relighting, we first consider decomposing hard shadows from scenes to avoid ambiguities with the albedo during material estimation. The traditional graphics method uses ray marching to calculate hard shadows according to the relative locations between the main light source and geometry. However, modeling shadows is a challenge for inverse rendering where light positions are unknown.

We observe that in HDR environments, the light intensity in areas around light sources and in non-light source areas can differ by several times or even by orders of magnitude. 
Therefore, we can set $\mu$ as the intensity threshold to distinguish between light and non-light source areas.
Here, we assume that if the incident light intensity from any direction at a spatial point is lower than $\mu$, the point will be considered in a hard shadow area, which means $\xi=0$.
The incident light intensity can be derived from the pre-trained HDR-radiance field (Sec.~\ref{sec:lighting}). Formally:
\begin{equation}
\xi = 
\begin{cases}
{1}, & \text{if } \Gamma_{max} (\textbf{x}) \geq \mu, \\
{0}, & \text{if } \Gamma_{max} (\textbf{x}) < \mu,
\end{cases}
\end{equation}

\noindent where $\Gamma_{max} (\textbf{x}) = \max_{i} {\Gamma_{i}(\textbf{x}, \bm{\omega}_i)}, \ i=1,2,...,N$ and \(\Gamma_{i}\) is incoming radiance intensity at spatial position \(\textbf{x}\) along \(\bm{\omega}_{i}\), which is the sum of radiance (along RGB channels) from the environment map \(E(\textbf{x},\bm{\omega}_i)\). \(\textit{N}\) is the number of sample rays. 
Since the radiance RGB values of each channel range from 0 to 1 for non-light source areas, we set \(\mu=3\) to distinguish light and non-light areas. 
Thus, this simple strategy can help to simulate hard shadows at unknown light locations. It not only takes into account the complexity of multiple light sources but also recovers shadows caused by intense indirect lighting.

To accelerate the process of acquiring hard shadows, we use the calculated $\xi$ as pseudo ground truth and firstly introduce a \textit{hard shadow field} $\Theta_{h}\textbf{(x)}$ to learn the hard shadow $S_{hard}$ at any input 3D location \(\textbf{x}\), which can efficiently model a continuous hard shadow distribution under unknown light locations.
However, with a limited number of sample rays, the hard shadows will include noises, especially at the edges of the shadows.
To address this issue, we additionally propose a differentiable soft shadow field in Sec.~\ref{sec:brdf}. The hard shadow field provides initial network parameters to the differentiable soft shadow field.

\subsection{BRDF}\label{sec:brdf}
Physically based rendering (PBR) provides a more accurate representation of how light interacts with material properties~\cite{pharr2023physically,karis2013real}. We adopt the physically-based microfacet BRDF model of Unreal Engine~\cite{karis2013real} to approximate the surface reflectance property and introduce a BRDF MLP \(\textit{f}_{r}\) to model the albedo $\hat{A}$ and roughness $\hat{R}$ of the scene. 
However, directly optimizing such a BRDF MLP can easily lead to non-convergence of roughness, due to the insufficient distribution of viewpoints in the training set as well as self-occlusion issues~\cite{invrender}. Moreover, shadows are prone to be inadvertently baked into the albedo~\cite{nvdiffrec} during direct optimization. We observe that the decomposition of albedo and shadow is more related to the diffuse component, while the specular component primarily influences the roughness. Hence, we propose a \textit{\textbf{three-stage}} material estimation strategy to alleviate this ambiguity problem.

Inspired by~\cite{zhu2022learning} and~\cite{nvdiffrec}, we leverage the Monte Carlo rendering algorithm and image-based lighting representation to recover albedo $\hat{A}$, roughness $\hat{R}$, and soft shadow \(S_{soft}\). The rendering equation (Eq.~\ref{con:renderequation}) can be written as:
\begin{align}
    L_{o}(\hat{\textbf{x}},\bm{\omega}_o) &= \int_{\Omega^+} E(\hat{\textbf{x}},\bm{\omega}_i) (\frac{\hat{A}}{\pi} + f_{s}(\hat{\textbf{x}}, \bm{\omega}_i,\bm{\omega}_o)) (\bm{\omega}_i \cdot \hat{\bm{n}}_x) \text{d}\bm{\omega}_i \label{eq:render} \\
    &= L_{o,d}(\hat{\textbf{x}}) + L_{o,s}(\hat{\textbf{x}},\bm{\omega}_o)\label{eq:render2}
\end{align}

\noindent where \(\hat{\bm{n}}_x\) is normal at surface point \(\hat{\textbf{x}}\), \(\bm{\omega}_i\) is light incident direction, \(\bm{\omega}_o\) is view direction.
According to the Lambertian model~\cite{kajiya1986}, albedo $\hat{A}$ contributes to the diffuse component. And roughness $\hat{R}$ affects the specular component \(f_{s}\) according to the Microfacet model. Therefore, we can represent the predicted rendering result $\hat{L}_o$ with a diffuse term $\hat{L}_{o,d}$ and a specular term $\hat{L}_{o,s}$ (see Eq.~\ref{eq:render2}). The specular $\hat{L}_{o,s}$ is calculated by Monte Carlo integration. More details can be found in Sec.~\ref{con:brdfequation} of the Appendix.

We optimize the parameters of the rendering equation in the following stages:



\textbf{Stage 1: Albedo initialization with $S_{hard}$.} 
\noindent With coarse albedo $\hat{A}$ predicted by $f_r$, we introduce a hard shadow term $\xi$ to the diffuse term $L_{o,d}$ as shown in Eq.~\ref{eq:diffuse}. By incorporating hard shadows, we can ensure that albedo takes on the inherent colors of the surfaces without shadow artifacts. 
\begin{equation}
    L_{o,d}(\hat{\textbf{x}}) = \frac{\hat{A}}{\pi} \xi \int_{\Omega^+} E(\hat{\textbf{x}},\bm{\omega}_i) (\bm{\omega}_i \cdot \hat{\bm{n}}_x) \text{d}\bm{\omega}_i
    \label{eq:diffuse}
\end{equation}

\noindent Note that we have pre-computed the \textit{hard shadow} and \textit{irradiance} terms using respective MLPs $\Theta_{h}$ and \(\Theta_I\), which can effectively minimize the number of unnecessary ray samples in the material estimation process, thereby enhancing computational efficiency. Therefore, we can re-write the predicted diffuse radiance as $\hat{L}_{o,d}=\frac{\hat{A}}{\pi}S_{hard}I$. In particular, the coarse albedo will only be optimized when $S_{hard}=1$, which means the point is not in a hard shaow area.

\textbf{Stage 2: Albedo refinement with differentiable $S_{soft}$.}
As mentioned in Sec.~\ref{sec:shadow}, using hard shadows will introduce noises to the rendering results. Therefore, we propose a differentiable soft shadow $S_{soft}$ using an MLP to represent the \textit{soft shadow field} $\Theta_s(\textbf{x})$, which is designed to provide a more nuanced representation of shadows, extending beyond the binary constraints of the hard shadow field. Crucially, the initial parameters of this soft shadow field $\Theta_s$ are inherited from the hard shadow field $\Theta_h$. We simply replace the hard shadow with the differentiable soft shadow, \ie, $\hat{L}_{o,d}=\frac{\hat{A}}{\pi}S_{soft}I$, and update the predicted soft shadow by optimizing the diffuse radiance. Besides, since albedo and shadows interact with each other, we add an instance-level regularizer $\mathcal{L}_{albedo}$ to the albedo. We assume that the luminance of the albedo is uniformly distributed at the instance level, showing no luminance variations. Formally:
\begin{equation}
    \mathcal{L}_{albedo} = \Big| \sum_{i=1}^{K} ( \hat{A} - \Phi_{inv}(\frac{\sum_{p} \Phi(\hat{A}) \odot M_{i}}{\sum_{p} M_{i}} )) \Big| \label{eq:loss_albedo}
\end{equation}
\noindent where $\Phi(\cdot)$ is an operator to obtain the \(\textit{V}\) value in HSV color domain by converting RGB-space to HSV-space, $\Phi_{inv}(\cdot)$ is an inverse operator to convert HSV-space to RGB-space~\cite{nvdiffrecmc}, \(M_{i}\) is the prior instance segmentation mask, \(K\) denotes the instance classes of the indoor scene, $p$ denotes the minibatch of 3D points, and \(\odot\) denotes an element-wise product.

\textbf{Stage 3: Roughness refinement.}
In the former two stages, the roughness $\hat{R}$ is directly predicted by the BRDF MLP $f_r$. Considering roughness has a similar instance-level assumption, we introduce a roughness smooth regularizer \(\mathcal{L}_{rough}\), which constrains the roughness on the same instance level to be similar.
\begin{equation}
    \mathcal{L}_{rough} = \Big| \sum_{i=1}^{K} ( \hat{R} - \frac{\sum_{p} \hat{R} \odot M_{i}}{\sum_{p} M_{i}} ) \Big|
\end{equation}

\subsection{Training}\label{sec:training}
We optimize the geometry, spatially varying illumination, shadow, and SVBRDF of the indoor scenes in \textit{\textbf{three phases}}: 1) training of geometry and HDR radiance field, 2) training of hard shadow and irradiance field, 3) training of BRDF with soft shadow field.

\textbf{Phase 1: Training of geometry and HDR radiance field.}
We train the geometry network $f_d$ and the HDR-radiance network $f_c$ in an end-to-end manner using the following loss~\cite{yariv2021}:
\begin{equation}
    \mathcal{L}_1 = \mathcal{L}_{recon} + \lambda_{eik}\mathcal{L}_{eik} + \lambda_{normal}\mathcal{L}_{normal},
\end{equation}

\noindent where \(\mathcal{L}_{recon}\) = \(\sum_{\textbf{r} \in R} ||\hat{C}(\textbf{r}) - C(\textbf{r})||_1\), $R$ denotes the set of rays samples in the minibatch, $\hat{C}$ is the volume rendering results in Eq.~\ref{volumerendering}, ${C}$ is the ground truth color. \(\mathcal{L}_{recon}\) use \(\ell_1\) to minimize the difference between each pixel’s predicted radiance and its ground-truth radiance. \(\mathcal{L}_{eik}\) is an eikonal term~\cite{yariv2021} to regularize the gradients of geometry network formulated as:

\begin{equation}
    \mathcal{L}_{eik} = \sum_{\textbf{x} \in \mathcal{X}} (\|\nabla d(\textbf{x})\|_2 - 1)^2,
    \label{eq:eikonal}
\end{equation}
\noindent where $\mathcal{X}$ is the minibatch of 3D points uniformly sampled in 3D space and nearby surface. \(\mathcal{L}_{normal}\) is a normal smooth loss~\cite{verbin2022ref}, which constrains the predicted normals $\bm{n}_i$ connected to the sampled density gradient normals $\bar{\bm{n}}_i$ along the ray: 
\begin{equation}
    \mathcal{L}_{normal} = \sum_{i} T_i \alpha_{i} \| \bm{n}_i - \bar{\bm{n}}_i \|^2
\end{equation}
\noindent where $T_i$ is the transmittance value and $\alpha_{i}$ is the alpha value of the $i$th sample along the ray, as defined in Eq.~\ref{volumerendering}.

\textbf{Phase 2: Training of hard shadow and irradiance field.}
In Sec.~\ref{sec:lighting} and Sec.~\ref{sec:shadow}, we can obtain the incoming diffuse lighting \(L_{i,d}\) and hard shadow \(\xi\), which serve as the ground truth for supervising the predicted irradiance $I$ and hard shadow $S_{hard}$ via \(\ell_1\) loss.
\begin{equation}
    \mathcal{L}_2 = ||L_{i,d} - {I}||_1 +  ||\xi - S_{hard}||_1  \label{eq:loss_2}
\end{equation}

\textbf{Phase 3: Training of BRDF and soft shadow field.}
During the optimization of intrinsic decomposition, we adopt a three-stage training scheme in Sec.~\ref{sec:brdf} to train BRDF and soft shadow field to avoid ambiguities between albedo and shadow. 
In stage 1, we optimize \(\mathcal{L}_{render}\) = \(||L_o - \hat{L}_o||_1\) to update the coarse albedo $\hat{A}$ with fixed hard shadow $S_{hard}$ and irradiance $I$. Then, in stage 2, we learn the differentiable soft shadow $S_{soft}$ and refine albedo by Eq.~\ref{eq:loss_3} with $\lambda_{albedo}=10^{-4}$ and $\lambda_{rough}=0$. Finally, we fix the above parameters in diffuse term and refine roughness values by Eq.~\ref{eq:loss_3} with $\lambda_{albedo}=0$ and $\lambda_{rough}=10^{-4}$.
\begin{equation}
    \mathcal{L}_{3} = \mathcal{L}_{render} + \lambda_{albedo}\mathcal{L}_{albedo} + \lambda_{rough}\mathcal{L}_{rough},  \label{eq:loss_3}
\end{equation}
More details of the rendering equation and implementation can refer to Sec.~\ref{con:brdfequation} and ~\ref{sec:implement_ours} in the Appendix.


\section{Experiments}

Our SIR pipeline is validated through experiments conducted on both synthetic and real-world indoor datasets.

\subsection{Indoor Datasets}
Owing to the absence of shadow effects within the existing HDR multi-view datasets for indoor scenes, we recognize the need for a more robust evaluation of our methodology. Consequently, we have developed two distinct inverse rendering datasets tailored specifically for indoor scene evaluation: \textbf{1)} The synthetic dataset is an extension of the DM-NeRF dataset~\cite{wang2022}, featuring six indoor scenes generated using Blender Cycles Render. It includes 300 posed images for training and 200 for testing. Each image has $400\times400$ pixels. We alter the locations and colors of objects within these scenes and enhance various object properties (\eg, albedo, roughness, shadow, instance mask, \etc).
\textbf{2)} Our real-world dataset includes two indoor scenes with elaborate materials and sophisticated lighting conditions. Utilizing a professional camera, we capture 120 HDR images for each scene, merging each HDR image from three distinct exposures (from $\frac{1}{15000}$ s to $\frac{1}{8}$ s). The captured images are processed to a resolution of $512\times512$ pixels, 85\% for training and 15\% for testing. These two datasets provide challenging indoor scenes for inverse rendering analysis.

\subsection{Implementation}
All neural fields in our network are implemented by multi-layer perceptrons (MLPs) with ReLU activations. For the geometry network $f_d$ and HDR-radiance network $f_c$, we follow the default setting from VolSDF~\cite{yariv2021}, where $f_d$ is an 8-layer MLP with hidden dimension 256 and $f_c$ is a 4-layer MLP with hidden dimension 256. The architectures of irradiance MLP, shadow MLPs, and BRDF MLP all contain 4 layers with 256 hidden units. Positional encoding~\cite{mildenhall2021} is applied to the input 3D locations and directions with 10 and 4 frequency components, respectively. For the SDF-based neural radiance field, we implement the two networks in PyTorch and optimize using Adam~\cite{kingma2014adam} with a learning rate of  \(5 \times 10^{-4}\) for 250K iterations. For irradiance and hard shadow estimation, we use the Adam optimizer with a learning rate of \(5 \times 10^{-4}\) for 10K iterations, and we use 512 sample rays at any surface point to compute the irradiance (Sec.~\ref{sec:lighting}) and shadow (Sec.~\ref{sec:shadow}). In all three stages of material estimation, we use the Adam optimizer with a learning rate of \( 10^{-3}\) for 25K iterations, and we use 128 samples to compute the BRDF term (Sec.~\ref{sec:brdf}). All experiments are conducted on a single NVIDIA GeForce RTX 3090 GPU. The entire training process takes approximately 40 hours and is divided into three phases: 16 hours for the first phase, 8 hours for the second phase, and 16 hours for the third phase.

\subsection{Inverse Rendering}
We select four notable and widely recognized inverse rendering methods as our baselines: 1) NVDIFFREC~\cite{nvdiffrec}, 2) IBL-NeRF~\cite{iblnerf}, 3) PhySG~\cite{physg}, and 4) InvRender~\cite{invrender}. Specifically, NVDIFFREC employs a DMTet-based approach, utilizing a differentiable marching tetrahedral layer to learn object-level meshes. IBL-NeRF, on the other hand, is a NeRF-based method for inverse rendering. PhySG and InvRender use an SDF-based neural radiance field for inverse rendering, which are both evaluated on object-level datasets. 
For fair comparisons on indoor scenes, we have minor modifications to adapt these baselines. Specifically, we align the geometric backbone of both InvRender~\cite{invrender} and PhySG~\cite{physg} with our geometry representation (\ie, VolSDF~\cite{yariv2021}), considering that both use SDF-based neural radiance fields for geometry. Further details are provided in Sec.~\ref{sec:implement_baselines} in Appendix. 

\begin{figure*}[t]
\centering
  \includegraphics[width=0.95\linewidth]{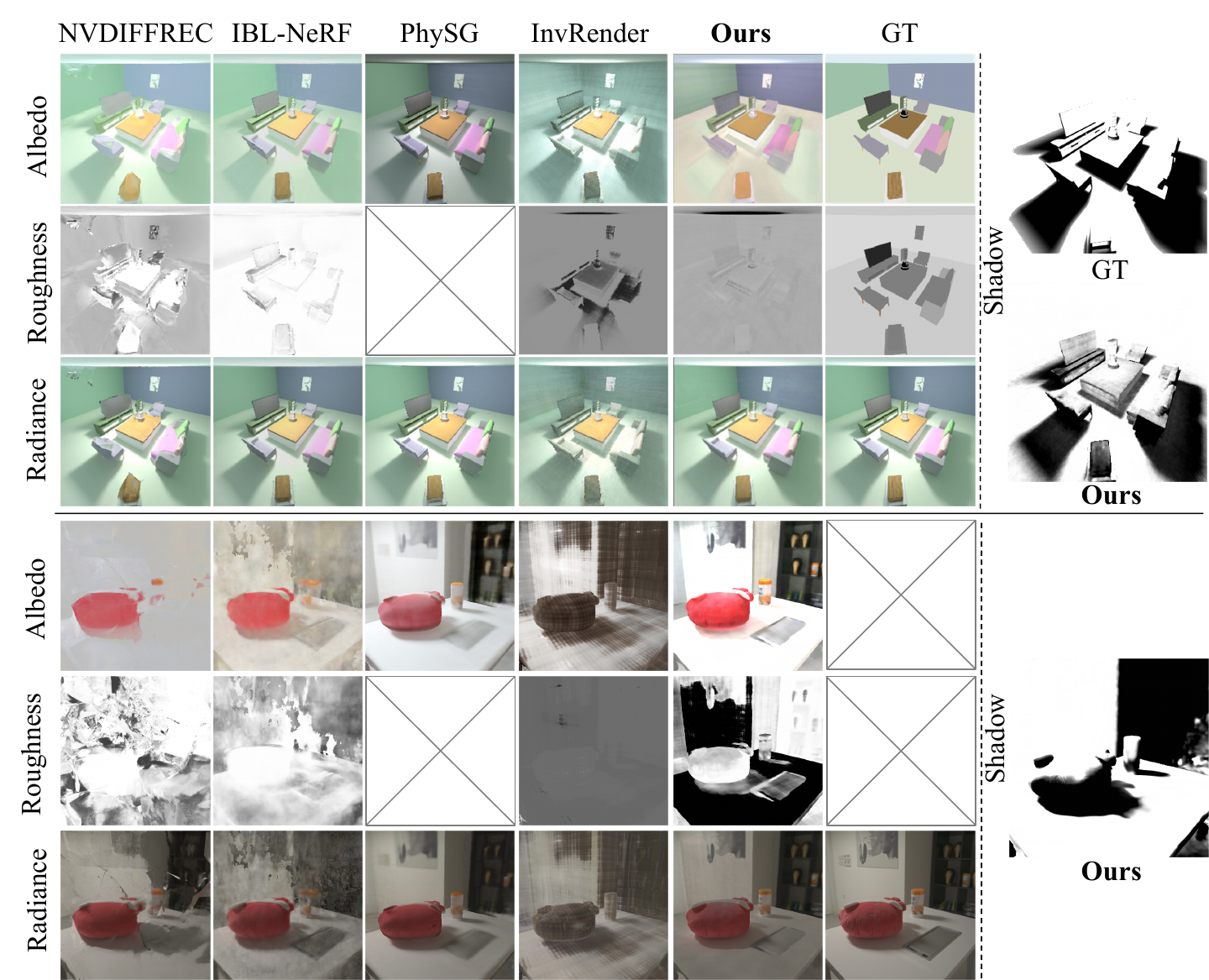}
  \vspace{-0.2cm}
\caption{Qualitative results of all methods on two datasets (\textit{Syn: Restroom, Real: Office}).}
\label{fig:exp_qualitative}
\end{figure*}

For the synthetic dataset, we report image quality metrics: PSNR, SSIM~\cite{karras2018progressive}, and LPIPS~\cite{zhang2018} on test viewpoints for view synthesis results and albedo, as well as the mean squared error (MSE) for roughness and shadow predictions. 
In particular, since our \nickname{} is the unique method that individually recovers shadows, we can only report the MSE on shadow predictions of our method. Besides, the ground truth shadows generated by the Blender renderer are binary, so we convert our soft shadows to binary for metric computation.
The evaluations of our method, as evidenced by the quantitative results in Table~\ref{tab:exp_syn} and qualitative results in Fig.~\ref{fig:exp_qualitative}, demonstrate that our \nickname{} outperforms existing methods in decomposing material attributes like albedo and roughness, and more notably, in shadow extraction. This leads to superior view synthesis results compared to other methods on indoor scenes. These results convincingly demonstrate the efficiency of the proposed shadow term in enhancing the overall view synthesis. 

Furthermore, we report view synthesis results of the real-world dataset. Our method exhibits a remarkable capability to handle complex lighting conditions in real-world indoor scenes in Table~\ref{tab:exp_realworld} and Fig.~\ref{fig:exp_qualitative}, which is crucial for achieving reliable material estimations for inverse rendering. 
The PSNR of our view synthesis results are higher than most of the baselines and slightly lower than IBL-NeRF. Meanwhile, the values of SSIM and LPIPS indicate that our results have less distortion and greater perceptual similarity to the targets.

\begin{table*}[t]  
\centering
\caption{Quantitative results for all baselines and ours. All scores are calculated as an average across 6 scenes from the synthetic dataset.}
\vspace{-0.3cm}
\resizebox{1.\linewidth}{!}{
  \tiny
  \begin{tabular}{l|ccc|c|c|ccc}
    \toprule
     &  \multicolumn{3}{c}{Albedo}  & \multicolumn{1}{c}{Roughness}  & \multicolumn{1}{c}{Shadow}  & \multicolumn{3}{c}{View Synthesis}\\
    & PSNR$\uparrow$  &  SSIM $\uparrow$  & LPIPS $\downarrow$ & MSE $\downarrow$ & MSE $\downarrow$ & PSNR$\uparrow$  &  SSIM $\uparrow$  & LPIPS $\downarrow$ 
 \\
    \midrule
    NVDIFFREC \cite{nvdiffrec} & 16.6377 & 0.7906 & 0.3736 & 0.0531 &  - & 23.8048 & 0.8606 & 0.1863  \\
    IBL-NeRF \cite{iblnerf} & 16.7773 & 0.8468 & 0.2161 & 0.0564 & - & 24.5258 & \textbf{0.9263} & \textbf{0.0844} \\
    PhySG~\cite{physg} & 10.5322 & 0.7076 & 0.3838 & - & - &  26.1667 & 0.9202 & 0.1014 \\
    InvRender~\cite{invrender} & 12.5401 & 0.6740 & 0.4935 & \textbf{0.0412} & - & 23.8582 & 0.7521 & 0.3923 \\
    \textbf{\nickname{} (Ours)} & \textbf{20.2767} & \textbf{0.8600} & \textbf{0.2154} & 0.0445 & \textbf{0.0541} & \textbf{28.5456} & 0.9258 & 0.0964 \\
    \bottomrule
  \end{tabular}
}
\label{tab:exp_syn}
\vspace{-0.2cm}
\end{table*}

\begin{table}[t]
\centering
\begin{minipage}{.48\linewidth}
  \centering
  \caption{Quantitative results on \textit{relighting} for all baselines and ours. All scores are calculated as an average across 6 scenes from the synthetic dataset.}
    \vspace{-0.3cm}
    \resizebox{1.\linewidth}{!}{
      \tiny
      \begin{tabular}{l|ccc}
        \toprule
        & PSNR$\uparrow$  &  SSIM $\uparrow$  & LPIPS $\downarrow$ 
     \\
        \midrule
        NVDIFFREC \cite{nvdiffrec} & 14.1443 & 0.7175 & 0.5041  \\
        PhySG~\cite{physg} & 13.5117 & 0.7716 & 0.2385 \\
        InvRender~\cite{invrender} & 14.3757 & 0.7240 & 0.3746 \\
        \textbf{\nickname{} (Ours)} & \textbf{21.4497} & \textbf{0.8215} & \textbf{0.2941} \\
        \bottomrule
      \end{tabular}
    }
    \label{tab:exp_syn_relight}
    \vspace{-0.4cm}
\end{minipage}%
\quad
\begin{minipage}{.48\linewidth}
  \centering
  \caption{Quantitative results on \textit{view synthesis} for all baselines and ours. All scores are calculated as an average across 2 scenes from the real-world dataset.}
\vspace{-0.3cm}
\resizebox{1.\linewidth}{!}{
  \tiny
  \begin{tabular}{l|ccc}
    \toprule
    & PSNR$\uparrow$  &  SSIM $\uparrow$  & LPIPS $\downarrow$ 
 \\
    \midrule
    NVDIFFREC \cite{nvdiffrec} & 22.7146 & 0.7798 & 0.4456  \\
    IBL-NeRF \cite{iblnerf} & 2\textbf{4.2781} & 0.7491 & 0.5225 \\
    PhySG~\cite{physg} & 22.1492 & 0.7796 & 0.4581 \\
    InvRender~\cite{invrender} & 20.1327 & 0.6694 & 0.6384 \\
    \textbf{\nickname{} (Ours)} & 22.8176 & \textbf{0.8345} & \textbf{0.2931} \\
    \bottomrule
  \end{tabular}
}
\label{tab:exp_realworld}
\vspace{-0.4cm}
\end{minipage}
\end{table}

\subsection{Ablation Study}\label{con:ablation}
The key innovation of our method is the capacity to effectively isolate shadows while ensuring that these shadows are not incorporated into the albedo under an indoor scene. Therefore, this section is dedicated to evaluating the impact of shadow terms, differentiable soft shadow, and albedo regularizer:

\noindent\textbf{(1) Removing shadow terms:} We remove both $S_{hard}$ and $S_{soft}$ in phases 2, 3 to demonstrate the necessity of introducing shadow terms in inverse rendering.

\noindent\textbf{(2) Removing soft shadow:} Without the differentiable shadow $S_{soft}$ in phase 3, the pipeline re-renders images using hard shadows $S_{hard}$.

\noindent\textbf{(3) Removing albedo regularizer:} This ablation is to evaluate the enhancement of introducing instance-level albedo consistency.

\begin{figure*}[t]
\centering
  \includegraphics[width=0.99\linewidth]{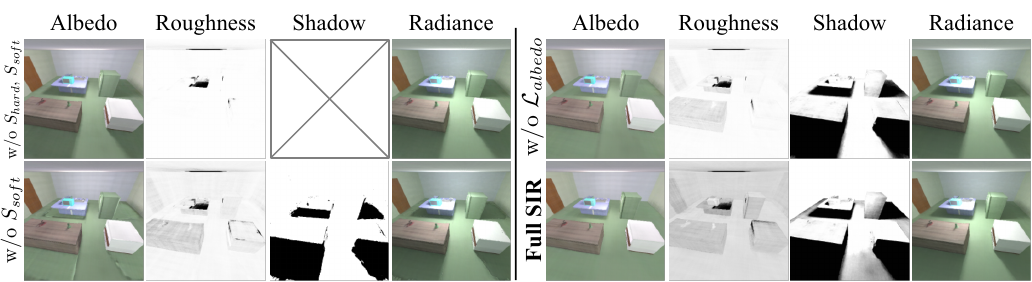}
  \vspace{-0.2cm}
\caption{Qualitative results of ablation study (\textit{Kitchen} in synthetic dataset).}
\label{fig:exp_ablation}
  \vspace{-0.2cm}
\end{figure*}

Table~\ref{tab:exp_abla} (1) clearly shows that removing the shadow term in the rendering equation, leads to a remarkable decline in the performance of novel view synthesis, adversely affecting the albedo estimation as well. The qualitative results in Fig.~\ref{fig:exp_ablation} further indicate that the absence of shadow terms results in incorrect albedo outputs with baked shadows, which then negatively impact the synthesis results. However, intuitively introducing a binary (hard) shadow still leads to suboptimal synthesis as shown in Table~\ref{tab:exp_abla} (2), particularly in the inaccuracies around shadow edges (see Fig.~\ref{fig:exp_ablation}). Therefore, learning a differentiable soft shadow after an initial hard shadow is necessary to preserve the shadow details in the synthesis images. Moreover, as demonstrated in Table~\ref{tab:exp_abla} (3), the integration of an instance-level albedo regularizer proves to be effective in maintaining consistency in albedo for each instance, thereby contributing to more accurate synthesis results.
Furthermore, the MSE values on roughness in Table~\ref{tab:exp_abla} clearly show that correctly decomposing shadows and albedo can prevent the tendency towards non-convergence in roughness, so as to recover accurate roughness values.

\begin{table*}[t] 
\centering
\caption{Quantitative results of ablation study on \nickname{}. All scores are calculated as an average across 6 scenes from the synthetic dataset.}
\vspace{-0.3cm}
\resizebox{1.\linewidth}{!}{
  \tiny
  \begin{tabular}{l|ccc|c|c|ccc}
    \toprule
     &  \multicolumn{3}{c}{Albedo}  & \multicolumn{1}{c}{Roughness}  & \multicolumn{1}{c}{Shadow}  & \multicolumn{3}{c}{View Synthesis}\\
    & PSNR$\uparrow$  &  SSIM $\uparrow$  & LPIPS $\downarrow$ & MSE $\downarrow$ & MSE $\downarrow$ & PSNR$\uparrow$  &  SSIM $\uparrow$  & LPIPS $\downarrow$ 
 \\
    \midrule
    (1) w/o $S_{hard}$, $S_{soft}$ & 16.6411 & 0.8084 & 0.2919 & 0.0712 & - & 28.2254  & 0.9221 & 0.0974  \\
    (2) w/o $S_{soft}$ & 19.3836 & 0.8052 & 0.3061 & 0.0627 & 0.0650 & 27.8201  & 0.8946 & 0.1897  \\
    (3) w/o $\mathcal{L}_{albedo}$ & 19.2597 & 0.8454 & 0.2406 & 0.0676 & 0.0863 & 28.3610 & \textbf{0.9258} & \textbf{0.0924}  \\
    \textbf{Full \nickname{}} & \textbf{20.2767} & \textbf{0.8600} & \textbf{0.2154} & \textbf{0.0445} & \textbf{0.0541} & \textbf{28.5456} & \textbf{0.9258} & 0.0964 \\
    \bottomrule
  \end{tabular}
}
\label{tab:exp_abla}
\vspace{-0.4cm}
\end{table*}

\begin{figure*}[t]
\centering
  \includegraphics[width=0.99\linewidth]{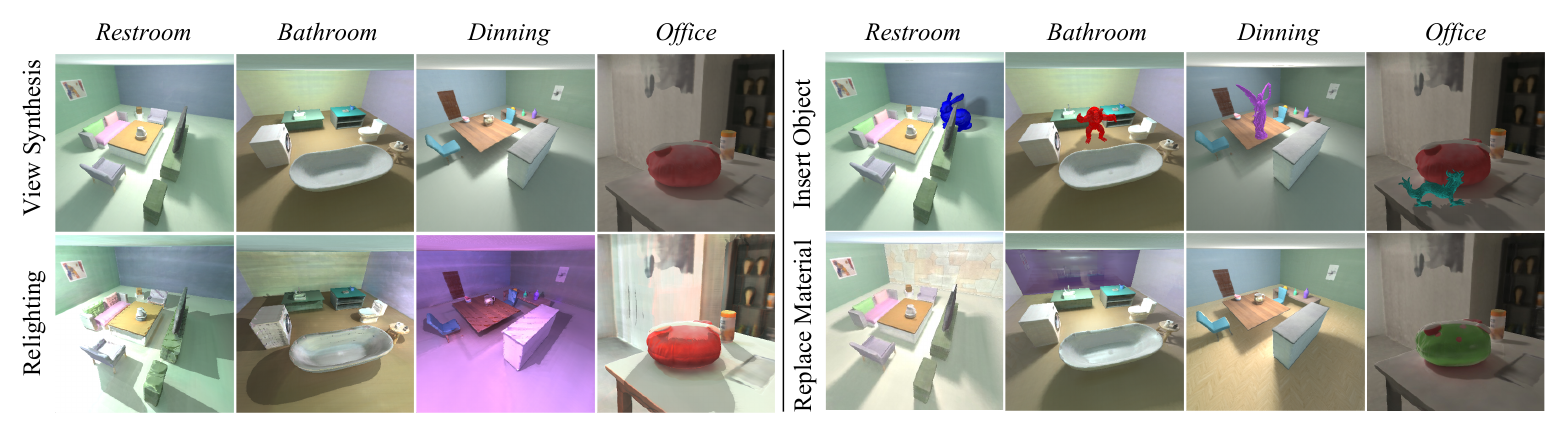}
  \vspace{-0.4cm}
\caption{Qualitative results of scene editing using our \nickname{} on synthetic dataset.}
\label{fig:exp_editing}
  \vspace{-0.4cm}
\end{figure*}

\subsection{Scene Editing}
Once our \nickname{} has effectively decomposed the intrinsic properties of indoor scenes, it enables the rendering of highly realistic novel view images of edited scenes by altering the values of these individual properties. To evaluate the superior performance of \nickname{} in accurately decomposing intrinsic elements, we have developed the following three distinct editing applications: 1) scene relighting, 2) object insertion, and 3) material replacement.

For quantitative evaluation on relighting of the synthetic dataset, we report the image quality metrics on test viewpoints as shown in Table~\ref{tab:exp_syn_relight}. Due to the limited capability of the irradiance module of IBL-NeRF in relighting, we can only compare the relighting performances with the other three baselines. Although other baselines have attempted to regularize albedo, they fail to account for the shadow term, which means albedo remains entangled with most of the shadows, significantly degrading the relighting quality. InvRender~\cite{invrender}, which accounts for visibility due to self-occlusion, only captures a minor portion of shadows generated by light source occlusion, and thus fails to significantly improve results.

In Fig.~\ref{fig:exp_editing}, we present the results of editing varied indoor scenes, encompassing both synthetic and real-world scenarios. The initial row at left shows the synthesized view images from our test set, serving as the source images for subsequent edits. The second row at left illustrates the outputs of free-viewpoint relighting experiments, featuring varied light positions and colors. Thanks to the cleanly decomposed albedo, free from any pre-existing shadows, our relighting experiments exhibit new realistic distributions of shadows. The first row at right is dedicated to object insertion, where we have incorporated four models from the Stanford 3D models~\cite{curless1996} into these scenes in diverse positions with different sizes and colors to compare the effect on re-rendering and shadow calculation. Our method achieves realistic shadowing and shading effects for these insertions. Furthermore, the second row at right shows material replacement: we randomly altered materials within the scenes, experimenting with different textures and levels of roughness (\eg, the wall in the \textit{Bathroom} scene). These convincingly realistic editing results demonstrate the effectiveness and robustness of our \nickname{} in decomposing the intrinsic properties of indoor scenes, which is fundamental for accurate inverse rendering applications.

\section{Conclusion}
We propose \nickname{}, a multi-view inverse rendering method on indoor scenes, effectively addressing the challenges of material and lighting decomposition by explicitly isolating shadows. Leveraging posed HDR images and an SDF-based radiance field, \nickname{} significantly enhances realism in material estimation and scene editing, surpassing previous methods with accurate shadow estimation. The innovative incorporation of shadows with a three-stage material estimation process substantially improves the quality of SVBRDFs. Extensive evaluations on diverse indoor datasets demonstrate the superiority of \nickname{} in both quantitative and qualitative aspects over existing methods. 

\noindent\textbf{Limitation}. First, our shadow and material estimation rely on fine geometry as an input. Second, for simplification, we neglect to refine the position of light sources for material estimation. In the future, we are interested in expanding our method to address the above limitations.

\bibliographystyle{splncs04}
\bibliography{main}

\clearpage

\appendix

\section{BRDF Model}\label{con:brdfequation}
In Sec~\ref{sec:brdf}, we adapt the microfacet BRDF model~\cite{walter2007microfacet,karis2013real} to approximate the surface reflectance property with a set of decomposed intrinsic terms, which can encapsulate the interaction of light with a surface. The microfacet model provides a robust framework for simulating the reflection from rough surfaces. The BRDF $f_r(\hat{\textbf{x}},\bm{\omega}_i,\bm{\omega}_o)$ is the ratio between the incoming and outgoing radiance. It is a function of the surface location $\hat{\textbf{x}}$, incoming direction $\bm{\omega}_i$, and the outgoing direction $\bm{\omega}_o$: 
\begin{equation}
    f_r(\hat{\textbf{x}},\bm{\omega}_i,\bm{\omega}_o) = f_d(\hat{\textbf{x}}) + f_s(\hat{\textbf{x}},\bm{\omega}_i,\bm{\omega}_o)
\end{equation}
\noindent where the BRDF comprises diffuse reflection $f_d$ and specular reflection $f_s$. In the first stage of BRDF estimation, we use the diffuse reflection term $f_d(\hat{\textbf{x}}) = \frac{\hat{A}_x}{\pi}$ based on a simple Lambertian model. Here, we introduce the specular BRDF $f_s$ which follows the Cook-Torrance model~\cite{cook1982reflectance}:
\begin{equation}
    f_s =  \frac{D(\bm{h}, \hat{\bm{n}}, \hat{R}) \cdot F(\bm{\omega}_o, \bm{h}) \cdot G(\bm{\omega}_i, \bm{\omega}_o, \hat{\bm{n}}, \hat{R})}{4(\hat{\bm{n}} \cdot \bm{\omega}_o)(\hat{\bm{n}} \cdot \bm{\omega}_i)}
\end{equation}
\noindent where $\hat{\bm{n}}$ is the surface normal at $\hat{\textbf{x}}$, $\hat{R}$ is roughness term, $\bm{h}$ is a half vector, $D$ denotes Normal Distribution Function (NDF), $F$ denotes Fresnel function and $G$ is the Geometry Factor.

The normal distribution function $D$ describes the probability density of microfacet orientations aligned with the half-vector $\bm{h}$. It is dependent on the macroscopic normal   $\bm{h}$ and the surface roughness $\hat{R}$, influencing the distribution of specular highlights, as follows: 
\begin{equation}
    D(\bm{h},\hat{\bm{n}},\hat{R}) = \frac{\alpha^2}{\pi ((\hat{\bm{n}} \cdot \bm{h})^2 (\alpha^2 - 1) + 1)^2}
\end{equation}
\noindent where the half-vector $\bm{h}$ is the normalized vector halfway between the incident light direction and the reflection direction, $\hat{\bm{n}}$ is the macroscopic surface normal, and $\alpha=\hat{R}^2$ is the surface roughness parameter, ranging from [0,1]. As $\alpha$ approaches 0, the surface tends towards a mirror-like reflection, while $\alpha$ approaching 1 indicates an extremely rough surface.

In the microfacet BRDF model, the Fresnel reflectance term \( F \) characterizes the proportion of light reflected from a material's surface at varying angles of incidence. This proportion changes with the angle of incidence, being lowest at normal incidence and highest at grazing angles for non-metallic materials. The Fresnel term can be efficiently computed using Schlick's approximation~\cite{schlick1994inexpensive}:
\begin{equation}
    F(\bm{\omega}_o, \bm{h}) = F_0 + (1 - F_0) \cdot (1 - (\bm{\omega}_o  \cdot \bm{h}))^5
\end{equation}
\noindent where $F_0$ is the reflectance at normal incidence, also known as the base reflectance and $\hat{\bm{\omega}_o} \cdot \bm{h}$ is the dot product between the outgoing radiance and the half-vector, dictating the cosine of the angle of incidence. In our method, we assume dielectric materials with $F_0 = 0.04$ in the Fresnel term.

In the realm of Physically Based Rendering (PBR), particularly within the context of the GGX Specular reflection model, the Geometry Function \( G \) assumes a pivotal role. It is primarily responsible for simulating the occlusion and shadowing effects attributable to the microsurface structure, crucial for accurately rendering light reflections on rough surfaces. The geometry function is typically computed using a combination of the Smith geometry function and the Schlick-GGX approximation. This formula is written as:
\begin{equation}
    G(\hat{\bm{n}}, \bm{\omega}_o, \bm{\omega}_i, \alpha) = G_1(\hat{\bm{n}}, \bm{\omega}_o, \alpha) \cdot G_1(\hat{\bm{n}}, \bm{\omega}_i, \alpha)
\end{equation}

\noindent Here, \( G_1 \) represents the monodirectional shadowing function:
\begin{equation}
    G_1(\hat{\bm{n}}, \bm{\omega}_o, \alpha) = \frac{\hat{\bm{n}} \cdot \bm{\omega}_o}{(\hat{\bm{n}} \cdot \bm{\omega}_o)(1 - k) + k}
\end{equation}
\begin{equation}
    G_1(\hat{\bm{n}}, \bm{\omega}_i, \alpha) = \frac{\hat{\bm{n}} \cdot \bm{\omega}_i}{(\hat{\bm{n}} \cdot \bm{\omega}_i)(1 - k) + k}
\end{equation}

\noindent where $\alpha$ is the parameter indicating surface roughness, $k$ is a parameter derived from the roughness, typically calculated as  $k = \frac{\alpha^2}{2}$.

\section{Additional Implementation Details}

\subsection{Implemetaions of SIR}\label{sec:implement_ours}

In Sec.~\ref{sec:training}, we use a three-phase training strategy to implement our framework, which includes the geometry and HDR-radiance field, irradiance and hard shadow field, and material and soft shadow estimation. The details are as follows:

\textbf{Phase 1: Geometry and HDR radiance field.}
We jointly optimize the geometry network $f_d$ and HDR-radiance network $f_c$ in this stage following VolSDF~\cite{yariv2021}.  During the training process, we observe that the vanilla VolSDF method falls short in effectively reconstructing scene geometry, a shortfall that considerably hampers the enhancement of shadows and materials. Consequently, we explore the integration of supplementary constraints to address these limitations. In synthetic datasets, we enhance the VolSDF to add an extra output for predicting normals. We supervise the predicted normals $\bm{n}$ using gradient-based normal $\bar{\bm{n}}$, resulting in smoother predicted normals. In real-world datasets, the geometric performance of VolSDF significantly declines compared to synthetic datasets. Therefore, inspired by the MonoSDF~\cite{yu2022monosdf}, we integrate ground truth normal to constrain our gradient-based normal $\bar{\bm{n}}$. and the ground truth normal is predicted by~\cite{eftekhar2021omnidata}. We optimize our geometry and HDR-radiance network for 250K iterations with a batch size of 1024 in this stage, which takes about 15 hours for a scene. 

\textbf{Phase 2: Hard shadow and irradiance field.}
For the representation of irradiance $I$ and hard shadow $S_{hard}$, we use two separate MLPs, each consisting of 4 layers of 256 hidden units with a rectified linear unit (ReLU) activation function. In addition, we encode the input surface position $\hat{\textbf{x}}$ with 10 levels of periodic functions, respectively, before feeding them into our network. 
We use the incoming diffuse lighting \(L_{i,d}\) in Sec.~\ref{sec:lighting} and hard shadow \(\xi\) in Sec.~\ref{sec:shadow} as the pseudo ground truths to supervise the predicted irradiance $I$ and hard shadow $S_{hard}$ via \(\ell_1\) loss. We use 512 sample directions over hemisphere \(\Omega^+\) to precompute the diffuse lighting \(L_{i,d}\) and hard shadow \(\xi\) at sampled surface points. To reduce the number of samples, we set the threshold $\mu$ as 4 in synthetic scenes and 3 in real-world scenes to obtain the hard shadow \(\xi\) by distinguishing HDR radiance intensity between the main light and environment light. We optimize irradiance and hard shadow MLPs for 10K iterations with a batch size of 256 in this stage, which takes about 7 hours for a scene. 

\textbf{Phase 3: BRDF and soft shadow field.}
In the section of material estimation (Sec.~\ref{sec:brdf}), the BRDF MLPs include four fully connected layers, each with 512 hidden units and ReLU activation functions. Following these four layers, the BRDF feature network is divided into separate albedo and roughness layers with 512 hidden units and Sigmoid activation. Specifically, albedo layer outputs the albedo term $\hat{A} \in \mathbb{R}^{3} $ and roughness layer outputs the roughness term $\hat{R} \in \mathbb{R}^{1}$ at surface position \(\hat{\textbf{x}}\). We use 64 sample directions (8 $\times$ 16) over hemisphere \(\Omega^+\) at sampled surface points for the specular environment map, and then when we use the specular environment map as specular incoming lighting with specular BRDF term $f_s$ to compute the rendering equation. We use important sampling to sample the incoming light radiance by interpolating the specular environment map based on Nvdiffrast~\cite{laine2020modular}, the important sampling number is 128.

For the training strategy of material estimation, we first freeze the final roughness layer and shadow term from the hard shadow MLP $\Theta_{h}$ and optimize the BRDF layer and albedo layer based on the diffuse component $L_{o,d}$ to obtain a coarse albedo with residual shadow. Next, we optimize the soft shadow MLP $\Theta_{s}$ and the albedo while keeping the roughness layer fixed. Here, the initial network parameters of the soft shadow MLP are inherited from the hard shadow MLP, providing a reasonable initial value to facilitate faster network convergence. Finally, we use instance-level albedo and roughness regularization while keeping the soft shadow MLP parameters fixed. For real-world datasets, we obtain the instance mask using the SegAny model~\cite{kirillov2023segment}. In this stage, we optimize our BRDF model for 25K iterations with a batch size of 256, which takes approximately 25 hours for a single scene.
We set weights \(\lambda_{eik} = 0.1\), \(\lambda_{normal} = 0.001\), \(\lambda_{albedo} = 0.0003\), \(\lambda_{rough} = 0.001\) in our experiments.

\subsection{Implemetaions of Baselines}\label{sec:implement_baselines}
As shown in Table~\ref{tab:exp_syn} and Fig.~\ref{fig:exp_qualitative}, we selected four prominent and widely acknowledged inverse rendering methods as our baselines: 1) NVDIFFREC~\cite{nvdiffrec}, 2) IBL-NeRF~\cite{iblnerf}, 3) PhySG~\cite{physg}, and 4) InvRender~\cite{invrender}. While TexIR~\cite{textir}, a recent notable approach for indoor inverse rendering using multi-view images, was also considered. However, due to its focus on large indoor scenes, which is incompatible with our dataset, it was excluded from our experimental baseline. Additionally, NeRFactor~\cite{zhang2021nerfactor}, a pioneering work in multi-view neural inverse rendering, was not included as a baseline in our study, given its comprehensive prior evaluation in InvRender~\cite{invrender}. 
Considering that our method incorporates additional normal constraints to improve geometric reconstruction, we make specific adjustments to ensure fairness in comparison. For synthetic datasets, we introduce a normal smooth loss, while for real-world scene datasets, we include normal ground truth supervision. To maintain consistency, we modify the geometric modules of PhySG~\cite{physg} and InvRender~\cite{invrender}, both of which are also based on SDF-based methods, to align with our approach. However, due to challenges in adapting IBL-NeRF~\cite{iblnerf} and NVDIFFREC~\cite{nvdiffrec} to match our geometric framework, we provide normal ground truth as geometric supervision for these methods.

\section{Details of Experiments}


\subsection{Additional Results on Synthetic Dataset}
\begin{figure*}[t]
\centering
  \includegraphics[width=0.9\linewidth]{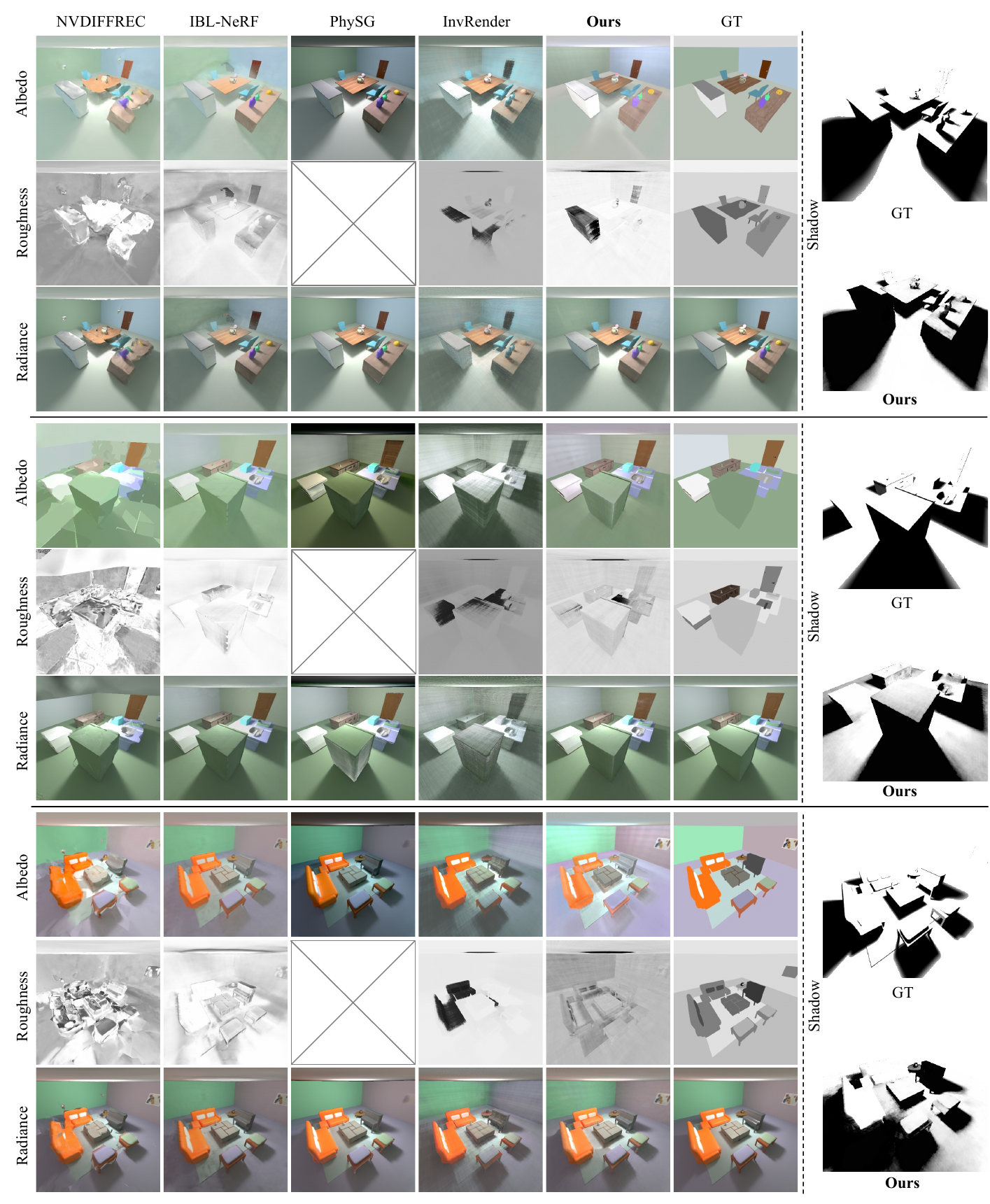}
  \vspace{-0.4cm}
\caption{Qualitative results of all methods on the synthetic dataset (\textit{Dinning}, \textit{Kitchen}, and \textit{Reception}).}
\label{fig:exp_qualitative_syn1}
  \vspace{-0.4cm}
\end{figure*}

\begin{figure*}[t]
\centering
  \includegraphics[width=1.\linewidth]{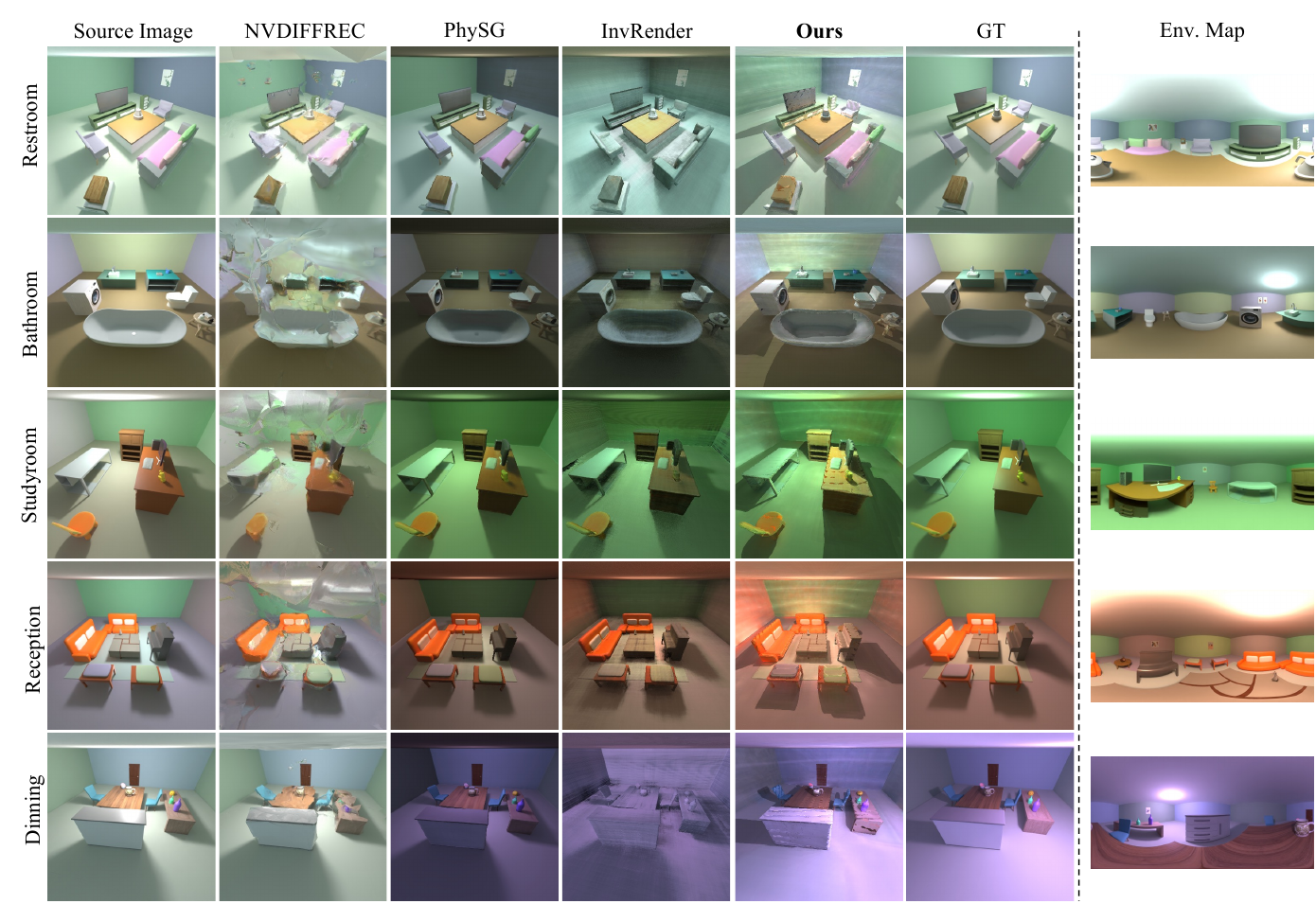}
  \vspace{-0.2cm}
\caption{Qualitative results on \textit{Relighting} of all methods on the synthetic dataset.}
\label{fig:exp_qualitative_relight_syn}
\end{figure*}

We present additional qualitative (see Fig.~\ref{fig:exp_qualitative_syn1}
and~\ref{fig:exp_qualitative_relight_syn}) and quantitative (see Table~\ref{tab:exp_syn_albedo_psnr}, ~\ref{tab:exp_syn_albedo_ssim},  ~\ref{tab:exp_syn_albedo_lpips}, ~\ref{tab:exp_syn_roughness}, ~\ref{tab:exp_syn_res_psnr}, ~\ref{tab:exp_syn_res_ssim}, and ~\ref{tab:exp_syn_res_lpips} comparison results on the synthetic dataset including 6 scenes. Compared to other inverse rendering methods, our approach excels in estimating albedo and decomposing shadows. Crucially, it effectively separates shadows and ensures that they are not incorrectly included in the albedo, especially under indoor lighting conditions. Note that PhySG~\cite{physg} and InvRender~\cite{invrender} directly use our trained SDF-based radiance field as their geometry backbone. Since the predicted roughness of PhySG is global, it cannot converge on our indoor scene dataset. Therefore, we ignore the comparison of the roughness MSE metric with PhySG.

\subsection{Additional Results on Real-World Dataset}
\begin{figure*}[t]
\centering
  \includegraphics[width=0.9\linewidth]{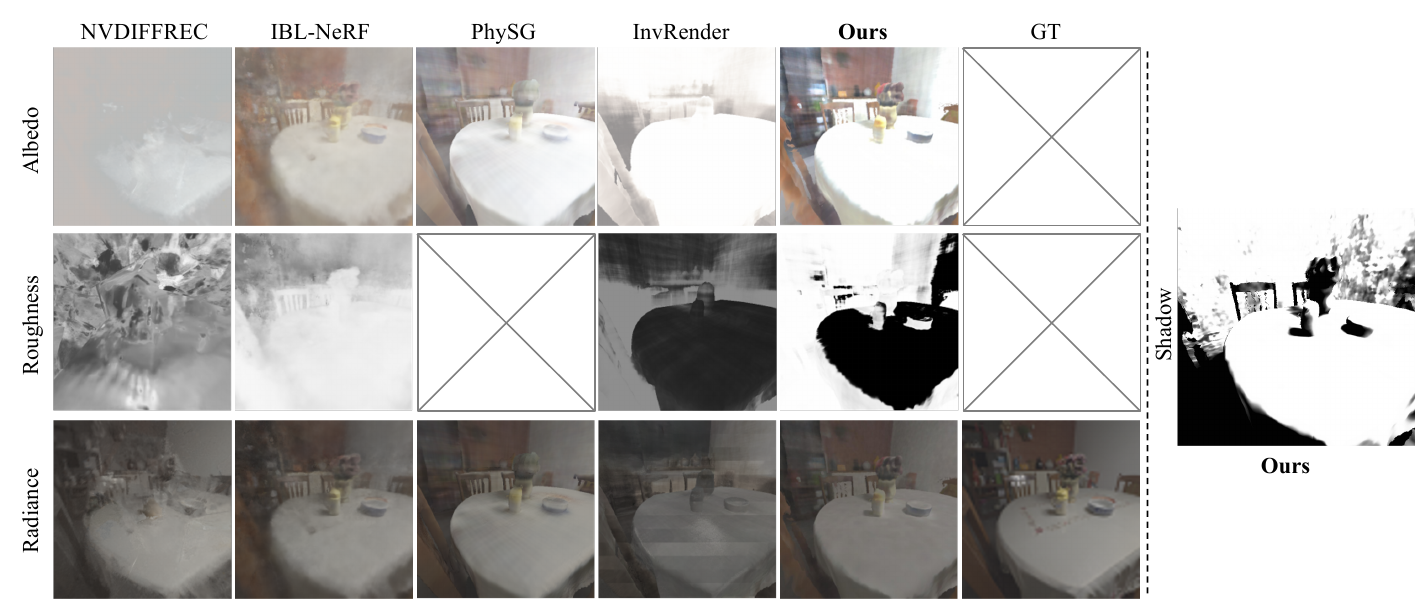}
  \vspace{-0.2cm}
\caption{Qualitative results of all methods on the real-world dataset (\textit{Dinning}).}
\label{fig:exp_realworld1}
\end{figure*}

Our comparative analysis on the real-world dataset, constrained to view synthesis metrics due to the unavailability of real scene material ground truths, is presented in Table~\ref{tab:exp_realworld1}. Although our method is slightly inferior to IBL-NeRF in terms of PSNR, it significantly outperforms other methods regarding SSIM and LPIPS metrics. Fig.~\ref{fig:exp_realworld1} provides additional qualitative results that highlight our exceptional decomposition ability. We effectively disentangle albedo and shadow components, achieving physical plausibility in our results.

\subsection{Additional Results for Ablation Studies}
\begin{figure*}[t]
\centering
  \includegraphics[width=1.\linewidth]{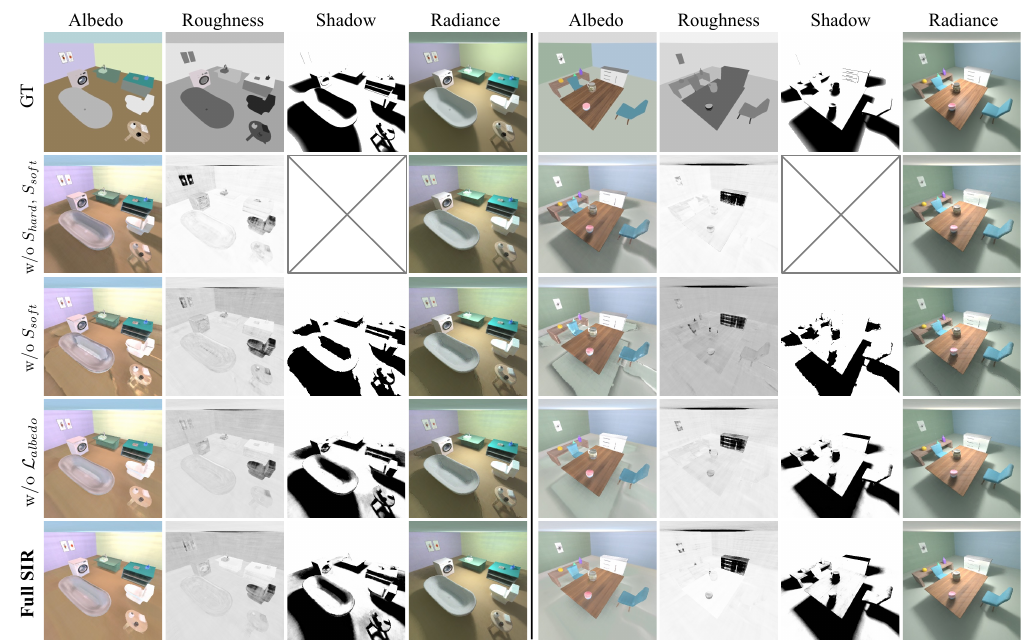}
  \vspace{-0.2cm}
\caption{Qualitative results of ablation study on the synthetic dataset (\textit{left: Bathroom, right: Dinning}).}
\label{fig:exp_ablation1}
\end{figure*}

We showcase the effectiveness of our material optimization on synthetic datasets in Table~\ref{tab:exp_abla_albedo_psnr},~\ref{tab:exp_abla_albedo_ssim},~\ref{tab:exp_abla_albedo_lpips},~\ref{tab:exp_abla_roughness},~\ref{tab:exp_abla_shadow},~\ref{tab:exp_abla_res_psnr},~\ref{tab:exp_abla_res_ssim}, and~\ref{tab:exp_abla_res_lpips}. As described in Sec.~\ref{con:ablation} in the main paper, Without shadow terms significantly impacts albedo estimation, leading to incorrect albedo outputs that erroneously incorporate baked shadows; Without the soft shadow term, residual shadows remain baked into the albedo, and due to the binary nature of hard shadows, they fail to accurately represent physically plausible shadow elements; Without the incorporation of an instance-level albedo regularizer, there would likely be a lack of illumination consistency in the albedo for individual instances, potentially leading to less accurate synthesis results. We show more qualitative results in Fig.~\ref{fig:exp_ablation1}.

\clearpage

\begin{table*}[t]
\centering
\caption{PSNR$\uparrow$ scores on \textit{Albedo} of all baselines and our method on 6 scenes of
synthetic dataset.}
\vspace{-0.4cm}
\resizebox{0.9\linewidth}{!}{
  \tiny
  \begin{tabular}{l|cccccc|c}
    \toprule
     & Bathroom & Dinning & Kitchen & Reception & Restroom & Study Room & Avg.
 \\
    \midrule
    NVDIFFREC \cite{nvdiffrec} & 16.5426 & 16.9962 & 16.9461 & 15.8834 & 15.8183 & 17.6397 & 16.6377 \\
    IBL-NeRF \cite{iblnerf} & 16.9009 & 17.4294  & 18.7986 & 14.9936 & 17.5475 & 14.9937 & 16.7773  \\
    PhySG~\cite{physg} & 11.9364 & 10.5226 & 9.1063 & 8.7889 & 10.7602 & 12.0790 &  10.5322 \\
    InvRender~\cite{invrender} & 14.5135 & 12.4979 & 11.6564 & 11.6064 & 13.7893 & 11.1773 &  12.5401 \\
    \textbf{\nickname{} (Ours)} & \textbf{20.5327} & \textbf{21.4604} & \textbf{20.0423} & \textbf{19.9779} & \textbf{21.3194} & \textbf{18.3272} &  \textbf{20.2767} \\
    \bottomrule
  \end{tabular}
}
\label{tab:exp_syn_albedo_psnr}
\vspace{-0.2cm}
\end{table*}

\begin{table*}[t]
\centering
\caption{SSIM$\uparrow$ scores on \textit{Albedo} of all baselines and our method on 6 scenes of
synthetic dataset.}
\vspace{-0.4cm}
\resizebox{0.9\linewidth}{!}{
  \tiny
  \begin{tabular}{l|cccccc|c}
    \toprule
     & Bathroom & Dinning & Kitchen & Reception & Restroom & Study Room & Avg.
 \\
    \midrule
    NVDIFFREC \cite{nvdiffrec} & 0.7809 & 0.7993 & 0.7630 & 0.7758 & 0.8046 & 0.8338 & 0.7906 \\
    IBL-NeRF \cite{iblnerf} & 0.8592 & 0.8425  & 0.8526 & \textbf{0.8349} & 0.8568 & 0.8349 & 0.8468  \\
    PhySG~\cite{physg} & 0.7472 & 0.7197  & 0.6582 & 0.6521 & 0.7261 & 0.7420 & 0.7076 \\
    InvRender~\cite{invrender} & 0.7253 & 0.7068  & 0.6145 & 0.7173 & 0.7124 & 0.5678 &  0.6740 \\
    \textbf{\nickname{} (Ours)} & \textbf{0.8654} & \textbf{0.8815}  & \textbf{0.8643} & 0.8293 & \textbf{0.8740} & \textbf{0.8452} & \textbf{0.8600} \\
    \bottomrule
  \end{tabular}
}
\label{tab:exp_syn_albedo_ssim}
\vspace{-0.2cm}
\end{table*}

\begin{table*}[t]
\centering
\caption{LPIPS$\downarrow$ scores on \textit{Albedo} of all baselines and our method on 6 scenes of
synthetic dataset.}
\vspace{-0.4cm}
\resizebox{0.9\linewidth}{!}{
  \tiny
  \begin{tabular}{l|cccccc|c}
    \toprule
     & Bathroom & Dinning & Kitchen & Reception & Restroom & Study Room & Avg.
 \\
    \midrule
    NVDIFFREC \cite{nvdiffrec} & 0.4326 & 0.3260 & 0.4615 & 0.3355 & 0.3657 & 0.3201 & 0.3736 \\
    IBL-NeRF \cite{iblnerf} & 0.2360 & 0.2451  & 0.2474 & 0.2545 & \textbf{0.2469} & 0.2545 & 0.2161  \\
    PhySG~\cite{physg} & 0.3275 & 0.3572  & 0.3953 & 0.4886 & 0.3820 & 0.3524 & 0.3838 \\
    InvRender~\cite{invrender} & 0.4089 & 0.4669  & 0.6359 & 0.4416 & 0.5083 & 0.4996 & 0.4935 \\
    \textbf{\nickname{} (Ours)} & \textbf{0.2064} & \textbf{0.1560}  & \textbf{0.2124} & \textbf{0.2155} & 0.2599 & \textbf{0.2420} & \textbf{0.2154} \\
    \bottomrule
  \end{tabular}
}
\label{tab:exp_syn_albedo_lpips}
\vspace{-0.1cm}
\end{table*}


\begin{table*}[t]\tabcolsep=0.1cm
\centering
\caption{MSE$\downarrow$ scores on \textit{Roughness} of all baselines and our method on 6 scenes of
synthetic dataset.}
\vspace{-0.4cm}
\resizebox{0.9\linewidth}{!}{
  \tiny
  \begin{tabular}{l|cccccc|c}
    \toprule
     & Bathroom & Dinning & Kitchen & Reception & Restroom & Study Room & Avg.
 \\
    \midrule
    NVDIFFREC \cite{nvdiffrec} & 0.0725 & 0.0620 & 0.0454 & 0.0450 & 0.0419 & 0.0515 & 0.0531 \\
    IBL-NeRF \cite{iblnerf} & 0.0824 & 0.0520  & \textbf{0.0334} & 0.0507 & 0.0691 & 0.0507 & 0.0564  \\
    PhySG~\cite{physg} & - & -  & - & - & - & - & - \\
    InvRender~\cite{invrender} & \textbf{0.0325} & \textbf{0.0240}  & 0.0373 & 0.0470 & 0.0691 & 0.0455 & \textbf{0.0412} \\
    \textbf{\nickname{} (Ours)} &  0.0547  & 0.0667 & 0.0408 & \textbf{0.0278} & \textbf{0.0318} & \textbf{0.0451} & 0.0445 \\
    \bottomrule
  \end{tabular}
}
\label{tab:exp_syn_roughness}
\vspace{-0.1cm}
\end{table*}


\begin{table*}[t]
\centering
\caption{PSNR$\uparrow$ scores on \textit{View Synthesis} of all baselines and our method on 6 scenes of synthetic dataset.}
\vspace{-0.4cm}
\resizebox{0.9\linewidth}{!}{
  \tiny
  \begin{tabular}{l|cccccc|c}
    \toprule
     & Bathroom & Dinning & Kitchen & Reception & Restroom & Study Room & Avg.
 \\
    \midrule
    NVDIFFREC \cite{nvdiffrec} & 25.0672 & 22.9778 & 21.7287 & 23.5271 & 24.0889 & 25.4392 & 23.8048 \\
    IBL-NeRF \cite{iblnerf} & 26.1249 & 24.2538  & 25.3530 & 23.1610 & 25.1013 & 23.1610 &  24.5258 \\
    PhySG~\cite{physg} & 28.5612 & 28.9428  & 18.9739 & 26.0657 & 28.7668 & 25.6900 & 26.1667 \\
    InvRender~\cite{invrender} & 26.1792 & 26.5549  & 13.9339 & 25.8879 & 25.8880 & 24.7052 & 23.8582 \\
    \textbf{\nickname{} (Ours)} & \textbf{28.9697} & \textbf{29.0003}  & \textbf{30.1561} & \textbf{26.7210} & \textbf{29.4958} & \textbf{26.9309} & \textbf{28.5456} \\
    \bottomrule
  \end{tabular}
}
\label{tab:exp_syn_res_psnr}
\vspace{-0.2cm}
\end{table*}

\begin{table*}[t]
\centering
\caption{SSIM$\uparrow$ scores on \textit{View Synthesis} of all baselines and our method on 6 scenes of synthetic dataset.}
\vspace{-0.4cm}
\resizebox{0.9\linewidth}{!}{
  \tiny
  \begin{tabular}{l|cccccc|c}
    \toprule
     & Bathroom & Dinning & Kitchen & Reception & Restroom & Study Room & Avg.
 \\
    \midrule
    NVDIFFREC \cite{nvdiffrec} & 0.8799 & 0.8340 & 0.8303 & 0.8530 & 0.9152 & 0.9059 & 0.8606 \\
    IBL-NeRF \cite{iblnerf} & 0.\textbf{9379} &  0.9072 & 0.8303 & \textbf{0.9273} & \textbf{0.9338} & 0.9273 &  \textbf{0.9263} \\
    PhySG~\cite{physg} & 0.9352 & \textbf{0.9383}  & 0.8676 & 0.9189 & 0.9336 & \textbf{0.9278} & 0.9202 \\
    InvRender~\cite{invrender} & 0.8054 & 0.7863  & 0.5798 & 0.8370 & 0.8370 & 0.7523 & 0.7521 \\
    \textbf{\nickname{} (Ours)} & 0.9276 & 0.9329  & \textbf{0.9267} & 0.9152 & 0.9279 & 0.9244 & 0.9258  \\
    \bottomrule
  \end{tabular}
}
\label{tab:exp_syn_res_ssim}
\vspace{-0.2cm}
\end{table*}

\begin{table*}[t]
\centering
\caption{LPIPS$\downarrow$ scores on \textit{View Synthesis} of all baselines and our method on 6 scenes of synthetic dataset.}
\vspace{-0.4cm}
\resizebox{0.9\linewidth}{!}{
  \tiny
  \begin{tabular}{l|cccccc|c}
    \toprule
     & Bathroom & Dinning & Kitchen & Reception & Restroom & Study Room & Avg.
 \\
    \midrule
    NVDIFFREC \cite{nvdiffrec} & 0.1771 & 0.2163 & 0.2391 & 0.1699 & 0.1832 & 0.1294 & 0.1863 \\
    IBL-NeRF \cite{iblnerf} & \textbf{0.0679} & 0.1250  & \textbf{0.0710} & \textbf{0.0790}  & \textbf{0.0592} & \textbf{0.0790} & \textbf{0.0844}  \\
    PhySG~\cite{physg} & 0.0756 & 0.0829  & 0.1760 & 0.1104 & 0.0770 & 0.0865 & 0.1014 \\
    InvRender~\cite{invrender} & 0.3047 &  0.3594 & 0.6263 & 0.2788 & 0.3022 & 0.3678 & 0.3923 \\
    \textbf{\nickname{} (Ours)} & 0.0916 & \textbf{0.0778}  & 0.0949 & 0.1044 & 0.1158 & 0.0940 & 0.0964  \\
    \bottomrule
  \end{tabular}
}
\label{tab:exp_syn_res_lpips}
\vspace{-0.2cm}
\end{table*}




\begin{table*}[t]
\centering
\caption{PSNR$\uparrow$ scores on \textit{Albedo} of ablation study on 6 scenes of
synthetic dataset.}
\vspace{-0.4cm}
\resizebox{0.9\linewidth}{!}{
  \tiny
  \begin{tabular}{l|cccccc|c}
    \toprule
     & Bathroom & Dinning & Kitchen & Reception & Restroom & Study Room & Avg.
 \\
    \midrule
    (1) w/o $S_{hard}$, $S_{soft}$ & 20.1723 & 17.7693 & 16.2957 & 16.9954 & 13.1248 & 15.4890 & 16.6411 \\
    (2) w/o $S_{soft}$ & 20.6228 & \textbf{22.0081}  & 18.7957 & 19.5625 & 16.7769 & \textbf{18.5355} &  19.3836 \\
    (3) w/o $\mathcal{L}_{albedo}$ & \textbf{20.7954} & 21.2082  & 19.5987 & 19.8224 & 16.6736 & 17.4599 & 19.2597 \\
    \textbf{Full \nickname{}} & 20.5327 &  21.4604 & \textbf{20.0423} & \textbf{19.9779} & \textbf{21.3194} & 18.3272 & \textbf{20.2767} \\
    \bottomrule
  \end{tabular}
}
\label{tab:exp_abla_albedo_psnr}
\vspace{-0.2cm}
\end{table*}

\begin{table*}[t]
\centering
\caption{SSIM$\uparrow$ scores on \textit{Albedo} of ablation study on 6 scenes of
synthetic dataset.}
\vspace{-0.4cm}
\resizebox{0.9\linewidth}{!}{
  \tiny
  \begin{tabular}{l|cccccc|c}
    \toprule
     & Bathroom & Dinning & Kitchen & Reception & Restroom & Study Room & Avg.
 \\
    \midrule
    (1) w/o $S_{hard}$, $S_{soft}$ & 0.8340 & 0.8314 & 0.8161 & 0.7935 & 0.7756 & 0.7998 & 0.8084 \\
    (2) w/o $S_{soft}$ & 0.8249 & 0.8310  & 0.8258 & 0.7925 & 0.7536 & 0.8035 & 0.8052  \\
    (3) w/o $\mathcal{L}_{albedo}$ & 0.8628 & 0.8713  & 0.8598 & 0.8183 & 0.8349 & 0.8255 & 0.8454 \\
    \textbf{Full \nickname{}} & \textbf{0.8654} & \textbf{0.8815}  & \textbf{0.8643} & \textbf{0.8293} & \textbf{0.8740} & \textbf{0.8452} & \textbf{0.8600} \\
    \bottomrule
  \end{tabular}
}
\label{tab:exp_abla_albedo_ssim}
\vspace{-0.2cm}
\end{table*}

\begin{table*}[t]
\centering
\caption{LPIPS$\downarrow$ scores on \textit{Albedo} of ablation study on 6 scenes of
synthetic dataset.}
\vspace{-0.4cm}
\resizebox{0.9\linewidth}{!}{
  \tiny
  \begin{tabular}{l|cccccc|c}
    \toprule
     & Bathroom & Dinning & Kitchen & Reception & Restroom & Study Room & Avg.
 \\
    \midrule
    (1) w/o $S_{hard}$, $S_{soft}$ & 0.2629 & 0.2541 & 0.2904 & 0.2649 & 0.3366 & 0.3423 & 0.2919 \\
    (2) w/o $S_{soft}$ & 0.2806 & 0.2650 & 0.2936 & 0.2717 & 0.3676 & 0.3580 & 0.3061  \\
    (3) w/o $\mathcal{L}_{albedo}$ & 0.2187 & 0.1888  & 0.2293 & 0.2324 & 0.2686 & 0.3059 & 0.2406 \\
    \textbf{Full \nickname{}} & \textbf{0.2064} &\textbf{ 0.1560}  & \textbf{0.2124} & \textbf{0.2155} & \textbf{0.2599} & \textbf{0.2420} & \textbf{0.2154} \\
    \bottomrule
  \end{tabular}
}
\label{tab:exp_abla_albedo_lpips}
\vspace{-0.1cm}
\end{table*}

\begin{table*}[t]
\centering
\caption{MSE$\downarrow$ scores on \textit{Roughness} of ablation study on 6 scenes of
synthetic dataset.}
\vspace{-0.4cm}
\resizebox{0.9\linewidth}{!}{
  \tiny
  \begin{tabular}{l|cccccc|c}
    \toprule
     & Bathroom & Dinning & Kitchen & Reception & Restroom & Study Room & Avg.
 \\
    \midrule
    (1) w/o $S_{hard}$, $S_{soft}$ & \textbf{0.0300} & 0.0719 & 0.0901 & 0.0343 & 0.0824 & 0.1187 & 0.2919 \\
    (2) w/o $S_{soft}$ & 0.0542 & \textbf{0.0247}  & 0.0613 & 0.0300 & 0.0863 & 0.1194 & 0.3061  \\
    (3) w/o $\mathcal{L}_{albedo}$ & 0.0526 & 0.0506  & 0.0736 & 0.0372 & 0.0836 & 0.1078 & 0.2406 \\
    \textbf{Full \nickname{}} & 0.0547 & 0.0667  & \textbf{0.0408} & \textbf{0.0278} & \textbf{0.0318} & \textbf{0.0451} & \textbf{0.2154} \\
    \bottomrule
  \end{tabular}
}
\label{tab:exp_abla_roughness}
\vspace{-0.1cm}
\end{table*}

\begin{table*}[t]
\centering
\caption{MSE$\downarrow$ scores on \textit{Shadow} of ablation study on 6 scenes of
synthetic dataset.}
\vspace{-0.4cm}
\resizebox{0.9\linewidth}{!}{
  \tiny
  \begin{tabular}{l|cccccc|c}
    \toprule
     & Bathroom & Dinning & Kitchen & Reception & Restroom & Study Room & Avg.
 \\
    \midrule
    (1) w/o $S_{hard}$, $S_{soft}$ & - & - & - & - & - & - & - \\
    (2) w/o $S_{soft}$ & 0.0513 & 0.0485  & 0.0503 & \textbf{0.0645} & 0.0752 & 0.1004 &  0.0650 \\
    (3) w/o $\mathcal{L}_{albedo}$ & 0.0599 & 0.0764  & 0.0667 & 0.1026 & 0.0768 & 0.1354 & 0.0863 \\
    \textbf{Full \nickname{}} & \textbf{0.0253} & \textbf{0.0475}  & \textbf{0.0350} & 0.1086 & \textbf{0.0520} & \textbf{0.0560} & \textbf{0.0541} \\
    \bottomrule
  \end{tabular}
}
\label{tab:exp_abla_shadow}
\vspace{-0.1cm}
\end{table*}

\begin{table*}[t]
\centering
\caption{PSNR$\uparrow$ scores on \textit{View Synthesis} of ablation study on 6 scenes of
synthetic dataset.}
\vspace{-0.4cm}
\resizebox{0.9\linewidth}{!}{
  \tiny
  \begin{tabular}{l|cccccc|c}
    \toprule
     & Bathroom & Dinning & Kitchen & Reception & Restroom & Study Room & Avg.
 \\
    \midrule
    (1) w/o $S_{hard}$, $S_{soft}$ & 28.4005 & 28.8354 & 28.6415 & 26.9213 & \textbf{29.6099} & 26.9436 & 28.2254 \\
    (2) w/o $S_{soft}$ & 28.4757 & 29.0484  & 29.2856 & 26.7187 & 26.9242 & 26.4678 & 27.8201  \\
    (3) w/o $\mathcal{L}_{albedo}$ & 28.8861 &  \textbf{29.5537} & 28.9922 & \textbf{26.9596} & 28.5293 & \textbf{27.2453} &  28.3610\\
    \textbf{Full \nickname{}} & \textbf{28.9697} & 29.0003  & \textbf{30.1561} & 26.7210 & 29.4958 & 26.9309 & \textbf{28.5456} \\
    \bottomrule
  \end{tabular}
}
\label{tab:exp_abla_res_psnr}
\vspace{-0.2cm}
\end{table*}

\begin{table*}[t]
\centering
\caption{SSIM$\uparrow$ scores on \textit{View Synthesis} of ablation study on 6 scenes of
synthetic dataset.}
\vspace{-0.4cm}
\resizebox{0.9\linewidth}{!}{
  \tiny
  \begin{tabular}{l|cccccc|c}
    \toprule
     & Bathroom & Dinning & Kitchen & Reception & Restroom & Study Room & Avg.
 \\
    \midrule
    (1) w/o $S_{hard}$, $S_{soft}$ & 0.9226 & 0.9288 & 0.9242 & 0.9160 & 0.9277 & 0.9134 & 0.9221 \\
    (2) w/o $S_{soft}$ & 0.9058 & 0.9070  & 0.9076 & 0.9013 & 0.8490 & 0.8968 &  0.8946 \\
    (3) w/o $\mathcal{L}_{albedo}$ & 0.9267 & 0.9322  & \textbf{0.9275} & 0.9145 & 0.9277 & \textbf{0.9260} & \textbf{0.9258} \\
    \textbf{Full \nickname{}} & \textbf{0.9276} & \textbf{0.9329}  & 0.9267 & \textbf{0.9152} & \textbf{0.9279} & 0.9244 & \textbf{0.9258} \\
    \bottomrule
  \end{tabular}
}
\label{tab:exp_abla_res_ssim}
\vspace{-0.2cm}
\end{table*}

\begin{table*}[t]
\centering
\caption{LPIPS$\downarrow$ scores on \textit{View Synthesis} of ablation study on 6 scenes of
synthetic dataset.}
\vspace{-0.4cm}
\resizebox{0.9\linewidth}{!}{
  \tiny
  \begin{tabular}{l|cccccc|c}
    \toprule
     & Bathroom & Dinning & Kitchen & Reception & Restroom & Study Room & Avg.
 \\
    \midrule
    (1) w/o $S_{hard}$, $S_{soft}$ & 0.0996 & 0.0828 & 0.0955 & 0.1049 & \textbf{0.0832} & 0.1185 & 0.0974 \\
    (2) w/o $S_{soft}$ & 0.1616 & 0.1839  & 0.1587 & 0.1656 & 0.2706 & 0.1980 & 0.1897  \\
    (3) w/o $\mathcal{L}_{albedo}$ & 0.0944 & 0.0811  & \textbf{0.0913} & 0.1089 & 0.0851 & \textbf{0.0936} & \textbf{0.0924} \\
    \textbf{Full \nickname{}} & \textbf{0.0916} & \textbf{0.0778}  & 0.0949 & \textbf{0.1044} & 0.1158 & 0.0940 & 0.0964 \\
    \bottomrule
  \end{tabular}
}
\label{tab:exp_abla_res_lpips}
\vspace{-0.2cm}
\end{table*}

\begin{table*}[h]
\centering
\caption{All scores on \textit{View Synthesis} of all baselines and our method on 2 scenes of real-world dataset.}
\vspace{-0.2cm}
\resizebox{1.\linewidth}{!}{
  \tiny
  \begin{tabular}{l|cc||c|cc||c|cc||c}
    \toprule
    &  \multicolumn{3}{c}{PSNR$\uparrow$}  & \multicolumn{3}{c}{SSIM$\uparrow$} & \multicolumn{3}{c}{LPIPS$\downarrow$}\\
     & Dinning & Office &  Avg. & Dinning & Office & Avg. & Dinning & Office & Avg. 
 \\
    \midrule
    NVDIFFREC \cite{nvdiffrec} & 21.3349 & \textbf{24.0942}  & 22.7146 & 0.7480  & 0.8116 & 0.7798 & 0.4686  & 0.4226 & 0.4456 \\
    IBL-NeRF \cite{iblnerf} & \textbf{26.234} & 22.3221  & \textbf{24.2781} & 0.7968 & 0.7013 & 0.7491 & 0.4285 & 0.6165 & 0.5225   \\
    PhySG~\cite{physg}  & 23.1329 & 21.1654  & 22.1492 & 0.7837 & 0.7755 & 0.7796 & 0.4871 & 0.4290 &  0.4581 \\
    InvRender~\cite{invrender}  & 21.3901 & 18.8752 & 20.1327 & 0.7025 & 0.6364 & 0.6694 & 0.6436 & 0.6331 & 0.6384  \\
    \textbf{\nickname{} (Ours)}  & 23.7331 & 21.9021  & 22.8176 & \textbf{0.8272} & \textbf{0.8418} & \textbf{0.8345} & \textbf{0.2854} & \textbf{0.3007} & \textbf{0.2931 } \\
    \bottomrule
  \end{tabular}
}
\label{tab:exp_realworld1}
\end{table*}

\end{document}